\documentclass{article}
\PassOptionsToPackage{numbers}{natbib}
\usepackage[preprint]{main}

\usepackage[utf8]{inputenc} % allow utf-8 input
\usepackage[T1]{fontenc}    % use 8-bit T1 fonts
\usepackage{hyperref}       % hyperlinks
\usepackage{url}            % simple URL typesetting

\usepackage[ruled]{algorithm2e}
\usepackage{amsmath,amssymb,amsfonts}
\usepackage{amsthm}
\usepackage{graphicx,subcaption}
\usepackage{wrapfig}

\usepackage{booktabs}       % professional-quality tables
\usepackage{amsfonts}       % blackboard math symbols
\usepackage{nicefrac}       % compact symbols for 1/2, etc.
\usepackage{microtype}      % microtypography
\usepackage{xcolor}         % colors

\title{A Resource-Adaptive Approach for Federated Learning under Resource-Constrained Environments}

\newtheorem{defMy}{Definition}
\newtheorem{thm}{\bf Theorem}
\newtheorem{remark}{Remark}
\newtheorem{cor}{Corollary}

\author{%
  Ruirui Zhang \\
  Shandong University\\
  Shandong, China \\
  \texttt{sherryz@mail.sdu.edu.cn} \\
  \And
  Xingze Wu \\
  Shandong University\\
  Shandong, China \\
  \texttt{wuxingze2022@163.com} \\
  \AND
  Yifei Zou \\
  Shandong University\\
  Shandong, China \\
  \texttt{yfzou@sdu.edu.cn} \\
  \And
  Zhenzhen Xie \\
  Shandong University\\
  Shandong, China \\
  \texttt{xiezz21@sdu.edu.cn} \\
  \And
  Peng Li \\
  the University of Aizu\\
  Japan \\
  \texttt{pengli@u-aizu.ac.jp} \\
  \And
  Xiuzhen Cheng \\
  Shandong University\\
  Shandong, China \\
  \texttt{xzcheng@sdu.edu.cn} \\
  \And
  Dongxiao Yu \\
  Shandong University\\
  Shandong, China \\
  \texttt{dxyu@sdu.edu.cn} \\
}

\begin{document}

\maketitle

\begin{abstract}
The paper studies a fundamental federated learning (FL) problem involving multiple clients with heterogeneous constrained resources. Compared with the numerous training parameters, the computing and communication resources of clients are insufficient for fast local training and real-time knowledge sharing. Besides, training on clients with heterogeneous resources may result in the straggler problem. To address these issues, we propose \textbf{Fed-RAA}: a \textbf{R}esource-\textbf{A}daptive \textbf{A}synchronous \textbf{Fed}erated learning algorithm. Different from vanilla FL methods, where all parameters are trained by each participating client regardless of resource diversity, Fed-RAA adaptively allocates fragments of the global model to clients based on their computing and communication capabilities. Each client then individually trains its assigned model fragment and asynchronously uploads the updated result. Theoretical analysis confirms the convergence of our approach. Additionally, we design an online greedy-based algorithm for fragment allocation in Fed-RAA, achieving fairness comparable to an offline strategy. We present numerical results on MNIST, CIFAR-10, and CIFAR-100, along with necessary comparisons and ablation studies, demonstrating the advantages of our work. To the best of our knowledge, this paper represents the first resource-adaptive asynchronous method for fragment-based FL with guaranteed theoretical convergence.
\end{abstract}

\section{Introduction}
Trillions of parameters enable the large-scale machine learning model to find the complex relationship hidden behind the inputs and outputs, and achieve a high performance on addressing the complex tasks in computer vision (CV) and natural language processing (NLP)~\cite{Artetxe_Bhosale_Goyal_Mihaylov_Ott_Shleifer_Lin_Du_Iyer_Pasunuru_et, Fedus_Zoph_Shazeer_2021, Lepikhin_Lee_Xu_Chen_Firat_Huang_Krikun_Shazeer_Chen_2020}. Whereas, it also poses great challenges in the training process, especially for those devices with limited computing and data resources~\cite{DBLP:journals/corr/abs-2403-19016,woisetschläger2023federated}. As an efficient distributed learning framework proposed by Google~\cite{17FL}, federated learning (FL) can effectively leverage the scattered resources among multiple clients in the learning process and maintain user privacy, as only the model information is exchanged, but not the data itself~\cite{PWNND24, FLL1}. Thus, it has been regarded as a classical approach for those resource-constrained clients to cooperatively obtain a comprehensive large-scale ML model~\cite{17FL}.

In the past decades, a series of relevant works have been proposed to train a large-scale ML model among resource-constrained clients under FL framework~\cite{Diao_Ding_Tarokh_2020,265005,9659532,SplitFed,RA-FED,woisetschläger2023federated,fine-tuning-LLM2023,Yuan_Wolfe_Dun_Tang_Kyrillidis_Jermaine_2022}, which generally fall into two categories. The works in the first branch~\cite{SplitFed,RA-FED,Yuan_Wolfe_Dun_Tang_Kyrillidis_Jermaine_2022} focus on clients with limited computational resources. By splitting the whole ML model into multiple submodels, each client only trains one of the submodels locally according to its computation ability. In the model aggregation step, those submodels will be aggregated into a global one and disseminated to the clients for the next round of training. 
In the second branch, the works mainly address the communication efficiency problem in FL since clients may have limited bandwidth. The techniques such as model compression, fine-tuning, and model scaling have been adopted in~\cite{265005},~\cite{woisetschläger2023federated,fine-tuning-LLM2023}, and~\cite{9659532} separately, which mitigate the communication overhead for clients in local model updating and global model disseminating steps. Additionally, in ~\cite{Diao_Ding_Tarokh_2020}, the computing and network resources constraints are simultaneously considered in their model splitting process. Whereas, all of the works mentioned above are executed under the synchronous aggregation mode, i.e. the model aggregation in each training round starts only when all the clients have their local models updated. Thus, the efficiency of the FL process will be delayed by the slowest client, which is often the client with the smallest computing and communication resources. The summary of the previous works is listed in Table~\ref{tab:algorithms}. To the best of our knowledge, few of the works consider the FL problem under the constrained computing and communication resources simultaneously under the asynchronous setting.

\begin{table}
    \vspace{-0.5em}
    \caption{Comparison of existing works}
    \label{tab:algorithms}
    \centering
    \vspace{\baselineskip}
    \begin{tabular}{llll}
        \toprule
        Related works & Computing Resource & Communication Resource & Aggregation mode  \\
        \midrule
 \cite{SplitFed}, \cite{RA-FED}, \cite{Yuan_Wolfe_Dun_Tang_Kyrillidis_Jermaine_2022}  & Limited & Sufficient & Synchronous \\
         \cite{fine-tuning-LLM2023}, \cite{FLL2} & Sufficient & Limited & Synchronous  \\
        \cite{Diao_Ding_Tarokh_2020}, \cite{pruning_2022} & Limited & Limited & Synchronous \\
        Our work & Limited & Limited & Asynchronous \\
        \bottomrule
    \end{tabular}    
    \vspace{-0.5em}
\end{table}
\begin{figure}[htb!]
    \centering
    \includegraphics[width=\linewidth]{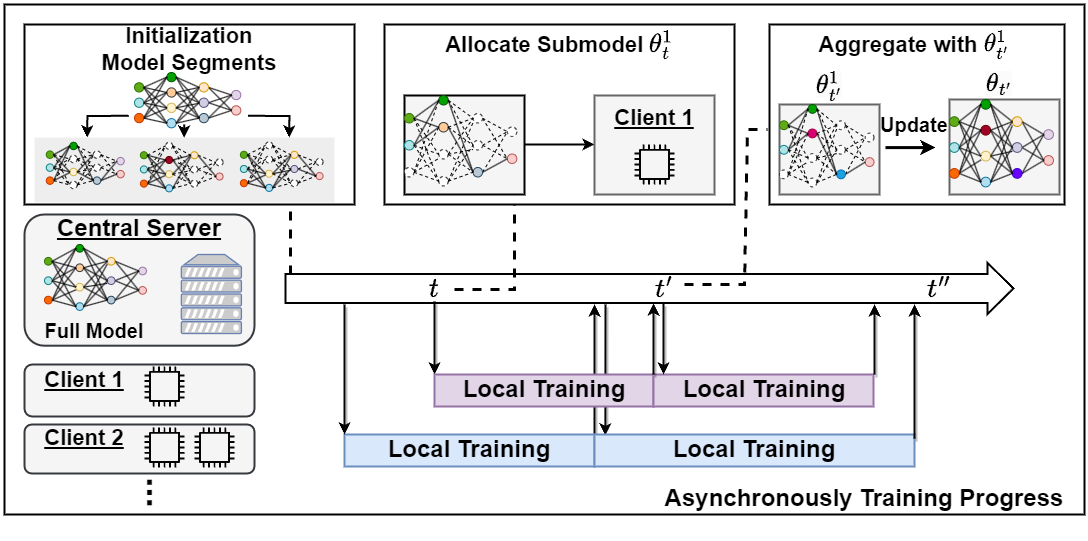}
    \caption{Overview of Fed-RAA}
    \label{fig:Motivate}
\end{figure}

To address the federated learning problem among multiple clients with heterogeneous constrained computing and communication resources, we propose a \textbf{R}esource-\textbf{A}daptive \textbf{A}synchronous \textbf{Fed}erated learning algorithm, named \textbf{Fed-RAA}. Since the local training, downloading, and uploading of an ML model with trillions of parameters are time-consuming or even not affordable for clients with limited resources, our method only designates a tailored submodel from the full model to each client for local training. For each client, once it receives a submodel from the parameter server, it trains the submodel based on its dataset. Considering that the synchronous FL process will be delayed by the slowest client in each training round, the asynchronous model aggregation is adopted in our Fed-RAA, i.e. once a client uploads its submodel, the parameters of the submodel will immediately updated to the global model. Theoretical analysis is presented to prove the convergence of our algorithm. Additionally, an online scheduling algorithm is designed as the designation strategy in our Fed-RAA, which considers the fairness of clients with respect to their computing and communication abilities and the fairness of parameters with respect to their opportunities to be updated. We also prove that our online algorithm has optimal performance on the submodel designation.

In summary, this paper introduces a \textbf{R}esource-\textbf{A}daptive \textbf{A}synchronous \textbf{Fed}erated learning algorithm, named \textbf{Fed-RAA}, to address the asynchronous FL problem among clients with heterogeneous constrained resources. Theoretical analysis is given to prove the convergence of Fed-RAA. Numerical results with necessary comparisons and ablation studies are presented, showing that our Fed-RAA has promising performance in Accuracy and Time cost. Additionally, an online scheduling algorithm is designed to designate the submodels to clients for local training, with a fairness guarantee.

\iffalse

The main contributions of this paper are summarized as follows:
\begin{itemize}
    \item We propose \textbf{Fed-RAA}, a novel \textbf{R}esource-\textbf{A}daptive \textbf{A}synchronous \textbf{Fed}erated learning algorithm tailored for training large ML models across clients with heterogeneous resources. In Fed-RAA, submodels from the global model are adaptively assigned to the clients 
    
    dynamically partitioned into submodels, where each client is adaptively allocated a submodel based on its resources, allowing asynchronous updates upon completion. This method not only addresses the resource diversity among clients but also mitigates the straggler problem effectively.
    
    \item We introduce an online greedy-based algorithm for dynamic model allocation in Fed-RAA, ensuring efficient resource utilization and fairness. We provide a detailed theoretical analysis demonstrating that our approach achieves convergence comparable to that of the optimal offline strategy. This breakthrough confirms the feasibility of the proposed method for practical, large-scale federated learning environments.
\end{itemize}
\fi

\section{Related Work}
The relevant works considering the FL problem of resource constrained clients can be divided into two categories. The works in the first category address the challenge of constrained computing resources. Yuan \textit{et al.} in \cite{Yuan_Wolfe_Dun_Tang_Kyrillidis_Jermaine_2022} decompose the original neural network into a collection of disjoint subnetworks, which are locally trained by clients and iteratively reassembled on the parameter server. SplitFed~\cite{SplitFed} splits the global model into the client-side model and server-side model. Each client just needs to train the client-side model, thus reducing its computing cost.  
In~\cite{RA-FED}, the submodel structures of clients are assigned according to their local computing resources. The works in the second category focus on the issue of limited communication resources in FL. Yang \textit{et al.} \cite{FLL2} 
introduce online model compression technique, to store model parameters in a compressed format and decompress them only when necessary. 
Moreover, DGC~\cite{lin2017deep} combines gradient sparsity and multiple optimization technologies to reduce communication costs with comparable accuracy. 
There are also some works considering both computing and communication resource constraints. For instance, HeteroFL~\cite{Diao_Ding_Tarokh_2020} decomposes the large model into multiple subnetworks, which are trained by the clients synchronously in the FL process. Simulations are presented in~\cite{Diao_Ding_Tarokh_2020} to validate the performance of HeteroFL.
Moreover, PruneFL~\cite{pruning_2022} adapts the model size during FL to reduce both communication and computation overhead and minimize the overall training time. 

However, the aforementioned works are built upon a synchronous FL framework. In this setup, the efficiency of FL in each training round is constrained by the speed of the slowest client, often referred to as the straggler. This synchronization may result in prolonged durations for each global training round, further leading to inefficient training. Different from the existing works, this paper investigates the asynchronous FL problem among multiple clients with heterogeneous constrained computing and communication resources, which is the novelty of our problem. 

\section{Preliminary}
We consider an FL system that consists of \(n\) clients and a parameter server \(S\), with reliable communication connections between the server and each client. 
Each client \(n\) (\(n=1,\ldots,N\)) has its own local dataset \(D^n\) with size \(\|D^n\|\) and limited computing resource \(cmp_n\). 
All clients train a global model in FL, the parameter set of which is denoted by \(\theta\). 

The existing works~\cite{Diao_Ding_Tarokh_2020,fedprun,Yuan_Wolfe_Dun_Tang_Kyrillidis_Jermaine_2022} have shown how to divide a global model into multiple submodels, each of which contains a fraction of the parameter \(\theta\). Thus, this paper no longer discusses the model division but starts from the assumption that the global model has already been divided into \(M\) submodels, denoted by \(\{\theta^1,\theta^2,\ldots,\theta^M\}\). 

The goal is to optimize the empirical risk minimization like traditional federated learning setting:
\begin{equation}\label{goal}
    \min_{\theta\in\mathbb{R}^d}{f(\theta)}:=\frac{1}{N}\sum_{n=1}^{N}{f_n(\theta)}
\end{equation}
where \(f_n(\theta) := \mathbb{E}_{\xi_n \sim D_n}[f_n(\theta,\xi_n)]\) is the local loss function of client \(n\) on dataset \(D_n\).

\section{Methodology}

\begin{figure}
    \centering
    \includegraphics[width=\linewidth]{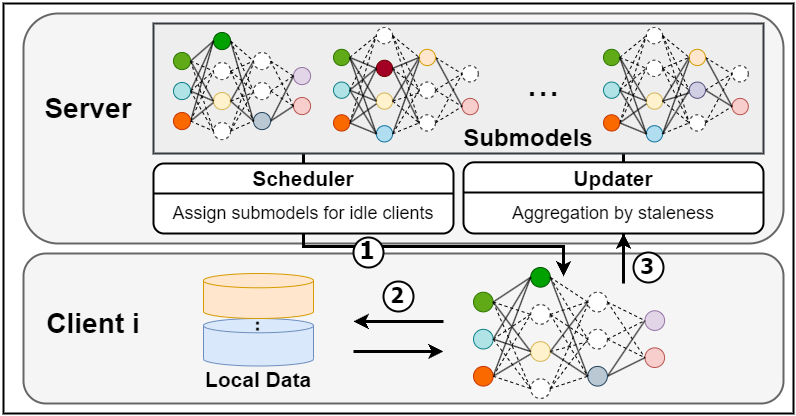} 
\caption{The pipeline of Fed-RAA}
\label{fig:framework}
\end{figure}

In this section, we design the resource-adaptive asynchronous FL algorithm, named Fed-RAA, in which the submodels \(\{\theta^1,\theta^2,\ldots,\theta^M\}\) are locally trained by the clients and asynchronously aggregated. 

\subsection{Algorithm Design}
The asynchronous training process takes \(t\) global epochs. 
In each epoch \(\tau\), the server assigns idle client \(n\) submodel \(\theta^{j}_{q(j)}\) for training, where \(j=h_\tau(n)\) is the index assigned to worker \(n\) in epoch \(\tau\) and \(q(j)\) denotes the number of update for submodel \(\theta^j\). 
Then in the \(t^{th}\) epoch, the server receives a locally trained model \(\theta_{new}^{j}\) from any arbitrary worker, and updates the corresponding submodel \(\theta^j\) by weighted averaging: \(\theta^{j}_{q(j)}\leftarrow(1-\alpha_t)\theta^{j}_{q(j)-1} + \alpha_t*\theta_{new}^{j}\), where \(\alpha\in (0,1)\) is the mixing hyperparameter. 

On an arbitrary device client \(n\), after receiving a submodel \(\theta^j_{q(j)}\) (potentially stale) from the server, we locally solve the following regularized optimization problem using SGD for several iterations: 
\begin{equation}\label{localGoal}
    \min_{\theta^j\in \mathbb{R}^d}{\mathbb{E}_{z^n\sim D^n}{f_n(\theta^j;z^n)+\frac{\rho}{2}{\|\theta^j-\theta^j_{q(j)}\|}}}
\end{equation}
For convenience, we define \(g_{\theta'}(\theta;z)=f(\theta;z)+\frac{\rho}{2}{\|\theta-\theta'\|^2}\). 

The server and clients conduct updates asynchronously, i.e., the server will immediately update the global model whenever it receives a locally trained submodel. 
Moreover, the communications between the server and clients are non-blocking. 
Thus, the server and clients can update the submodels at any time without synchronization, which is favorable when the devices have heterogeneous resources. 

The detailed algorithm is shown in Algorithm~\ref{alg1}. 
The submodel parameter \(\theta^j_{\tau,i}\) is updated in the \(i\)th local iteration after receiving \(\theta^j_{\tau}\) on device client \(n\). 
The data sample \(z^n_{\tau, i}\) is randomly drawn from local dataset \(D^n\) and \( I^n_t \) is the number of local iterations after receiving \(\theta^j_{\tau}\) on device client \(n\). 
Moreover, \(\gamma\) is the learning rate, and \(t\) is the total number of global epochs. 

As the pipeline is shown in Figure.~\ref{fig:framework}, for each client, the process begins with the server assigning an appropriate submodel via the scheduler module (Step 1). Upon receiving the submodel, each client initiates local training (Step 2). Once a client completes their local training, they upload their submodel to the server. The server then integrates the submodel into the global model using an updater (Step 3), thereby completing that round of updates. This continues for each client until the server signifies a halt.

\begin{algorithm}[htbp]

	\caption{Fed-RAA} 
 \label{alg1}
	\KwIn{Local Data \(D^n\) of client \(n\) on \(N\) clients.} 
        \SetAlgoLined
        \SetKwProg{Pro}{Progress}{:}{}
        \Pro{Server\((\alpha \in (0,1))\)}{
        Initialize \(\theta_0 = \{\theta_0^1,\theta_0^2,\ldots,\theta_0^M\}\), \(\alpha_t\leftarrow\alpha\), \(\forall t \in [T]\)\;
        Run Scheduler() thread and Updater() thread asynchronously in parallel\;
        }
        \SetKwProg{Thr}{Thread}{:}{}
        \Thr{Scheduler()}{
        Periodically trigger training tasks on some idle workers, and send \(\theta_{q(h_t(n))}^{h_t(n)}\) to worker \(n\) with time stamp\;
        Where \(h_t(n)\) is the index assigned to worker \(n\) in round \(t\)\;
        }
        \Thr{Updater()}{
        \For{\textit{epoch} \(t\in [T]\)}{
        Receive the pair \((\theta_{new}^{h_{\tau}(n)},\tau)\) from any worker\;
        Optional: \(\alpha_t\leftarrow \alpha * s(t-\tau)\), where \(s(\cdot)\) is a function of the staleness\;
        \(q(h_\tau(n)) \leftarrow q(h_\tau(n))+1\)\;
        \(\theta^{h_\tau(n)}_{q(h_\tau(n))}\leftarrow(1-\alpha_t)\theta^{h_\tau(n)}_{q(h_\tau(n))-1} + \alpha_t*\theta_{new}^{h_\tau(n)}\)\;
        }
        }
        \Pro{Worker()}{
        \For{\(n\in [N]\) in parallel}{
        \If{triggered by the scheduler}{
        Receive the pair of the global model and its time stamp \((\theta_{q(h_t(n))}^{h_t(n)},t)\) from the server\;
        \(j\leftarrow h_t(n)\) which is the index of model region\;
        \(\tau\leftarrow q(j)\)\;
        \(\theta_{\tau,0}^{j}\leftarrow\theta_{q(j)}^j\)\;
        Define \(g_{\theta_{q(j)}^j}(\theta^j;z)=f_n(\theta^j;z)+\frac{\rho}{2}\|\theta^j-\theta^j_{q(j)}\|^2\), where \(\rho > \mu\)\;
        \For{local iteration \(i\in [I^n_t]\)}{
        Randomly sample \(z^n_{\tau,i} \sim D^n\)\;
        Update \(\theta_{\tau, i}^j\leftarrow \theta_{\tau, i-1}^{j}-\gamma\nabla g_{\theta_{q(j)}^j}(\theta_{\tau, i-1}^j;z^n_{\tau,i})\)\;
        }
        Push \((\theta_{\tau,I^n_t}^j,t)\) to the server\;
        }
        }
        }
\end{algorithm}

\subsection{Convergence Analysis}
\begin{defMy}
\label{smooth}
    \textit{(Smoothness)} A differentiable function \( f \) is \( L \)-smooth if for \(\forall x, y, f(y) - f(x) \leq \langle \nabla f(x), y - x \rangle + \frac{L}{2} \| y - x \|^2\), where \( L > 0 \). 
\end{defMy}
\begin{defMy}
\label{wconv}
    \textit{(Weak convexity)} A differentiable function \( f \) is \( \mu \)-weakly convex if the function \( g \) with \( g(\theta) = f(\theta) + \frac{\mu}{2} \| \theta \|^2 \) is convex, where \( \mu \geq 0 \). 
    \( f \) is convex if \( \mu = 0 \), and non-convex if \( \mu > 0 \). 
\end{defMy}
\begin{thm}\label{thm1}
    Assume that every function \( F_n \) is \( L \)-smooth and \( \mu \)-weakly convex, and each worker executes at least \( I_{\text{min}} \) and at most \( I_{\text{max}} \) local updates before pushing models to the server. 
    We assume bounded delay \( t - \tau \leq K \). 
    The imbalance ratio of local updates is \( \delta = \frac{I_{\text{max}}}{I_{\text{min}}} \). 
    Furthermore, we assume that for \( \forall \theta \in \mathbb{R}^d, n \in [N] \), and \( \forall z \sim D^n \), we have \( \| \nabla f(\theta; z) \|^2 \leq V_1 \) and \( \| \nabla g_{\theta '}(\theta; z) \|^2 \leq V_2, \forall \theta ' \). 
    For any small constant \( \epsilon > 0 \), taking \( \rho \) large enough such that \( \rho > \mu \) and \( -(1 + 2\rho + \epsilon)V_2 + \rho^2 \| \theta_{\tau, i-1}^j - \theta_{\tau}^j \|^2 - \frac{\rho}{2} \| \theta_{\tau, i-1}^j - \theta_{\tau}^j \|^2 \geq 0, \forall \theta_{\tau, i-1}^j, \theta_{\tau}^j, j \in [M], \) and \( \gamma < \frac{1}{L} \), after \( Q \) updates for every submodel \( \theta^j \), Algorithm~\eqref{alg1} converges to a critical point: 
    \[ 
    \min_{q=0}^{T-1} \mathbb{E} \| \nabla F(\theta_{q}) \|^2 \leq \frac{\mathbb{E}[F(\theta_0) - F(\theta_T)]}{\alpha \gamma \epsilon T I_{\text{min}}} + \mathcal{O}\left(\frac{\gamma I_{\text{max}}^3 + \alpha K I_{\text{max}}}{\epsilon I_{\text{min}}}\right) + \mathcal{O}\left(\frac{\alpha^2 \gamma K^2 I_{\text{max}}^2 + \gamma K^2I_{\text{max}}^2}{\epsilon I_{\text{min}}}\right).
    \]
    Where \(T = Q*M\)
\end{thm}
Theorem\ref{thm1} shows the convergence rate of algorithm Fed-RAA by giving the upper bound on the gradient of all clients for all trained submodels. 
It is worth noting that in the convergence result, we ensure that all submodels can be updated uniformly, that is, \(T=Q*M\). 

\begin{remark}\label{rem2}
    Impact of the training delay bound \( K \). 
    Our convergence result shows that the smaller the upper bound \( K \) of staleness \( (t - \tau) \) would lead to a faster convergence rate and better performance in our federated learning process.
\end{remark}

\begin{cor}\label{cor1}
    Let all assumptions hold. Using \(\delta=\frac{I_{max}}{I_{min}}\), and taking \(\alpha = \frac{1}{\sqrt{I_{min}}}\), \(\gamma = \frac{1}{\sqrt{T}}\), \( I_{min} = T^{\frac{1}{5}}\), we have the convergence rate as follows: 
    \begin{align*}
    \min_{q=0}^{T-1}\mathbb{E}[\|\nabla F(\theta_q)\|^2] & \leq \mathcal{O}\left(\frac{1}{T^{\frac{3}{5}}}\right) + \mathcal{O}\left(\frac{\delta^3}{T^{\frac{1}{10}}}\right) + \mathcal{O}\left(\frac{K\delta}{T^{\frac{1}{10}}}\right) + \mathcal{O}\left(\frac{K^2\delta^2}{T^{\frac{1}{2}}}\right) + \mathcal{O}\left(\frac{K^2\delta^2}{T^{\frac{3}{10}}}\right).
    \end{align*}
\end{cor}
Corollary \ref{cor1} indicates that the term \(\mathcal{O}\left(\frac{\delta^3+K\delta}{T^{\frac{1}{10}}}+\frac{K^2\delta^2}{T^{\frac{3}{10}}}\right)\) will dominate the convergence rate.

The detailed theoretical proof of Theorem\ref{thm1} is provided in Supplement. 

\section{Online Submodel Assignment Algorithm}
As mentioned in Algorithm~\ref{alg1}, we rely on a submodel assignment algorithm to ensure an upper bound on the training delay, thereby facilitating the convergence analysis for the whole model.
The implementation of this algorithm can be either learning-based or scheduling-based. 
In this section, we propose an online \textbf{Gre}edy-based \textbf{R}esource \textbf{A}daptive submodel \textbf{A}ssignment algorithm, Gre-RAA, which is scheduling-based and proved as optimal as an offline solution on the upper bound of training delay \(K\). 

\subsection{Algorithm Design for Submodel Assignment }
Considering the varying communication resources between clients, we define \(com^{up}_{i,j}\) as the number of global epochs for client \(i\) to update submodel \(\theta^j\) and \(com^{d}_{i,j}\) as the download time. 
Assuming each client updates the same local iterations when receiving one region submodel, we define \(c_{i,j}=\frac{\|\theta^j\|\|D^n\|}{cmp_i} + com^{up}_{i,j} + com^{d}_{i,j}\) as the global epochs consumed by client \(i\) to train submodel \(\theta^j\). 

\textbf{Fairness.} Considering the limited heterogeneous computational and communication resources of users, we ensure 
the fairness of clients with respect to their computing and communication abilities
and the fairness of parameters with respect to their opportunities to be updated. Specifically, we strive to ensure that \textbf{(1)} if two clients receive their submodels at a similar time, the completion time for both should also be close, meaning that each client receives an adaptive submodel tailored to their resources, and \textbf{(2)} during the training process, each submodel has roughly equal opportunities or frequency of being updated.

To ensure such fairness, when a submodel is assigned by the server to the client \(n\) at the epoch \(t\), it must satisfy the following requirements: \textbf{(1)} the training delay of it for client \(n\) does not exceed the upper bound \( K \) and \textbf{(2)} it has been updated the fewest number of times so far. 

If multiple submodels meet the two aforementioned conditions, we randomly select one for assignment. Additionally, to ensure that every user has at least one submodel that satisfies the conditions for training, we can adjust the value of the hyperparameter \(K\). Specifically, we assume that the chosen \(K\) is sufficiently large so that the set of submodels meeting the first condition is not empty.

The details are shown in Algorithm~\ref{alg2}, which can be found in the Appendix.

\subsection{Algorithm Analysis}
\begin{thm}\label{opti}
    Given the asynchronous federated learning system proposed in Alg.~\ref{alg1}, there is an online scheduling-based model assignment algorithm, shown as {\tt Algorithm}~\ref{alg2}, reaching the same upper bound of training delay as that conducted by an optimal offline model assignment strategy. 
\end{thm}
The detailed theoretical proof of Theorem\ref{opti} is provided in Supplement. 

\section{Experiment}
To evaluate the performance of Fed-RAA and compare it with other algorithms, our experiments selected three common benchmark datasets in the field of image classification: CIFAR-10~\cite{CIFAR}, CIFAR-100~\cite{CIFAR}, and MNIST~\cite{MNIST}. In simulation, the convergence rate of the global model and its maximum achievable accuracy are observed. 
More details are provided in the Appendix. 
 
\subsection{Comparisons with Baselines}
To substantiate the superior performance of Fed-RAA, we benchmark it against state-of-the-art methods in FL, namely, \textbf{FedAvg}~\cite{FedAvg}, \textbf{FedSync}~\cite{fedsync}, \textbf{FedProx}~\cite{fedprox}, \textbf{FedPrun}~\cite{fedprun}, \textbf{SplitFed}~\cite{SplitFed}, and \textbf{RAM-Fed}~\cite{RA-FED}. 
FedAvg represents a classical synchronous full-model FL framework, while FedSync exemplifies the classical asynchronous approach. FedProx, SplitFed, and RAM-Fed focus primarily on the local computational limitations of users; in FedProx, users train the full model, whereas, in SplitFed and RAM-Fed, users employ a submodel for synchronous FL. FedPrun considers both computational and communication resources but operates within a synchronous FL framework. 

\begin{figure}[ht!]
    \centering 
    \begin{subfigure}{0.32\textwidth}
        \includegraphics[width=\linewidth]{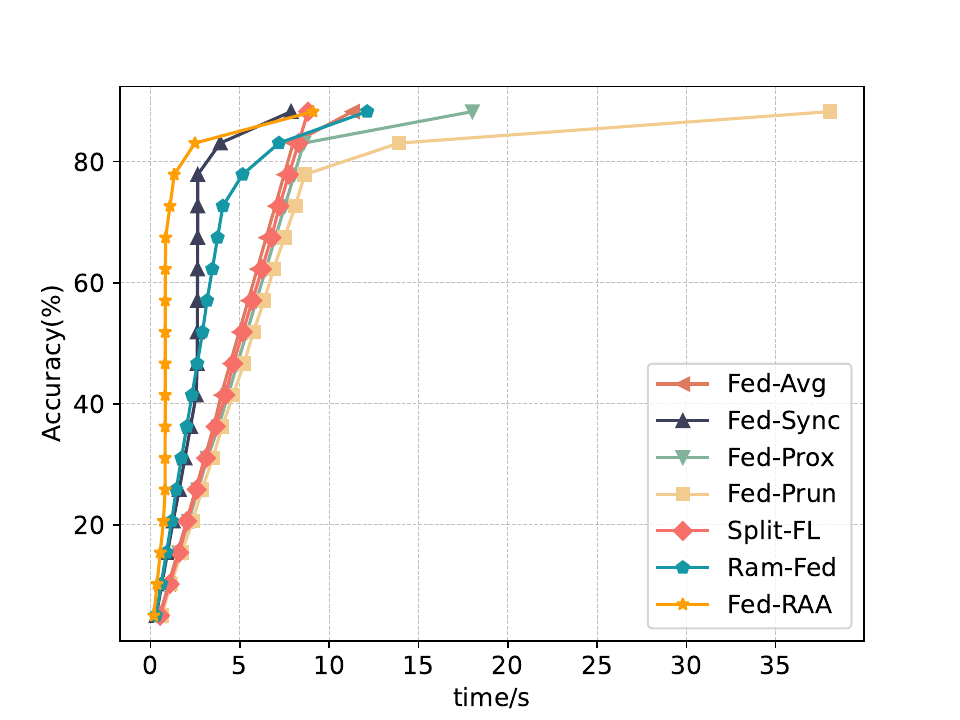}
        \caption{MNIST}
        \label{fig:mnist}
    \end{subfigure}
    \hfill
    \begin{subfigure}{0.32\textwidth}
        \includegraphics[width=\linewidth]{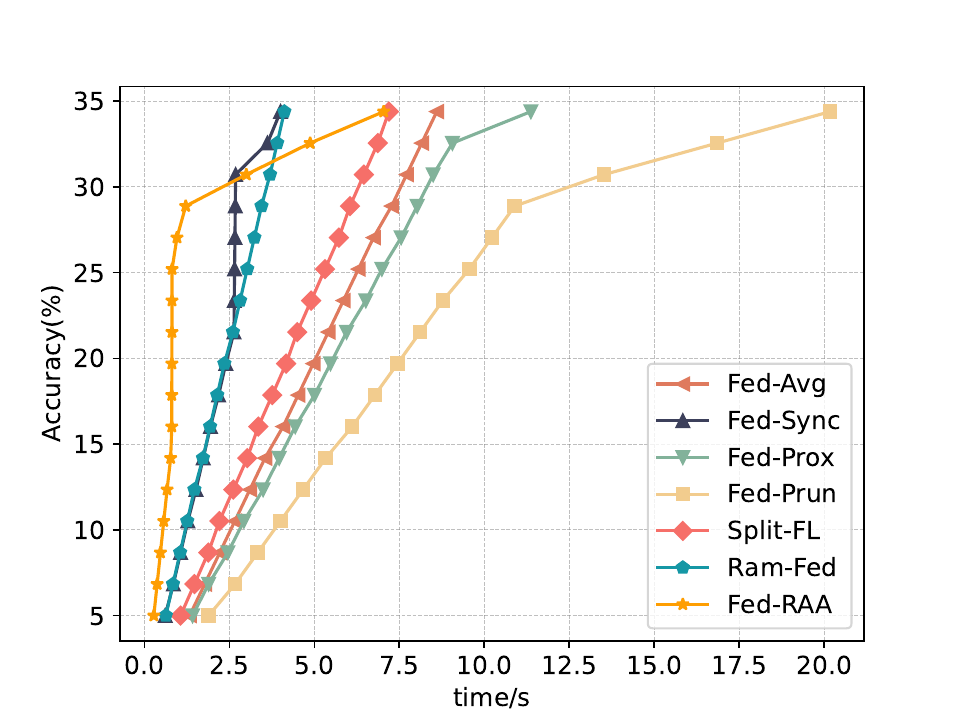}
        \caption{CIFAR-10}
        \label{fig:CIFAR-10}
    \end{subfigure}
    \begin{subfigure}{0.32\textwidth}
        \includegraphics[width=\linewidth]{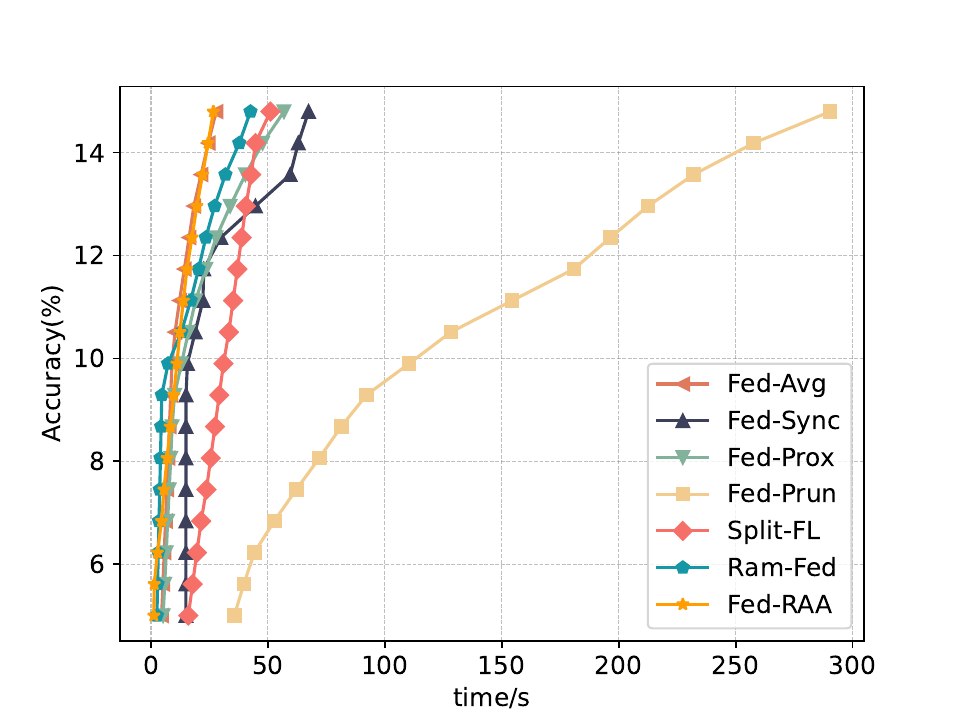}
        \caption{CIFAR-100}
        \label{fig:CIFAR-100}
    \end{subfigure}
    \hfill 
    \caption{Detailed results of comparison among different algorithms}
    \label{fig:exp1_detail}
\end{figure}

\begin{table}[ht!]
% \vspace{-0.5em}
\caption{Comparison of the time and epochs required to achieve specified accuracy.\\  <MNIST,0.93>, <CIFAR-10,0.45>, and <CIFAR-100,0.19> are the tuple of <Dataset, Accuracy>.}
\label{time_epoch_comparison}
  \centering
  \vspace{1em}
\begin{tabular}{lccc}
\toprule
 Time (s) & <MNIST,0.93> & <CIFAR-10,0.45> & <CIFAR-100,0.19> \\
\midrule
Fed-Avg  & 406.35  & 390.83  & 435.51  \\
Fed-Sync & 242.46  & 253.11  & 1123.25  \\
Fed-Prox & 866.40 & 870.44  & 910.98  \\
Fed-Prun & 776.82  & 975.48  & 3158.55  \\
Split-FL & 840.15  & 769.88  & 2181.59  \\
Ram-Fed  & 808.96  & 783.06  & 918.11  \\
Fed-RAA  & \textbf{91.06 } & \textbf{89.85} & \textbf{101.86} \\
\bottomrule
\end{tabular}
% \vspace{-0.5em}
\end{table}
We conducted experiments across various datasets with our algorithm and baseline algorithms and found that our approach converges more rapidly. As demonstrated in Table~\ref{time_epoch_comparison} and Fig.~\ref{fig:exp1_detail}, on the MNIST dataset, our algorithm achieved a 0.93\% accuracy in just 91.06 seconds. This faster convergence can be attributed to more frequent updates within shorter intervals due to its asynchronous nature. Similar effects were observed on other datasets as well. This highlights our algorithm's superiority in rapidly updating and converging compared to traditional synchronous methods and other asynchronous or submodel training approaches.

In the Appendix, we have included an additional table detailing accuracy and time costs. The findings are similar: Our Fed-RAA algorithm exhibits exceptional time efficiency across multiple datasets while sustaining competitive accuracy, making it an ideal option for federated learning scenarios that require swift model updates and deployments. It substantially surpasses traditional synchronous and several asynchronous methods, underscoring its utility in contexts where time efficiency and rapid convergence are paramount.
\subsection{Ablation Study}
\begin{figure}[ht!]
    \centering 
    \begin{subfigure}{0.32\textwidth}
        \includegraphics[width=\linewidth]{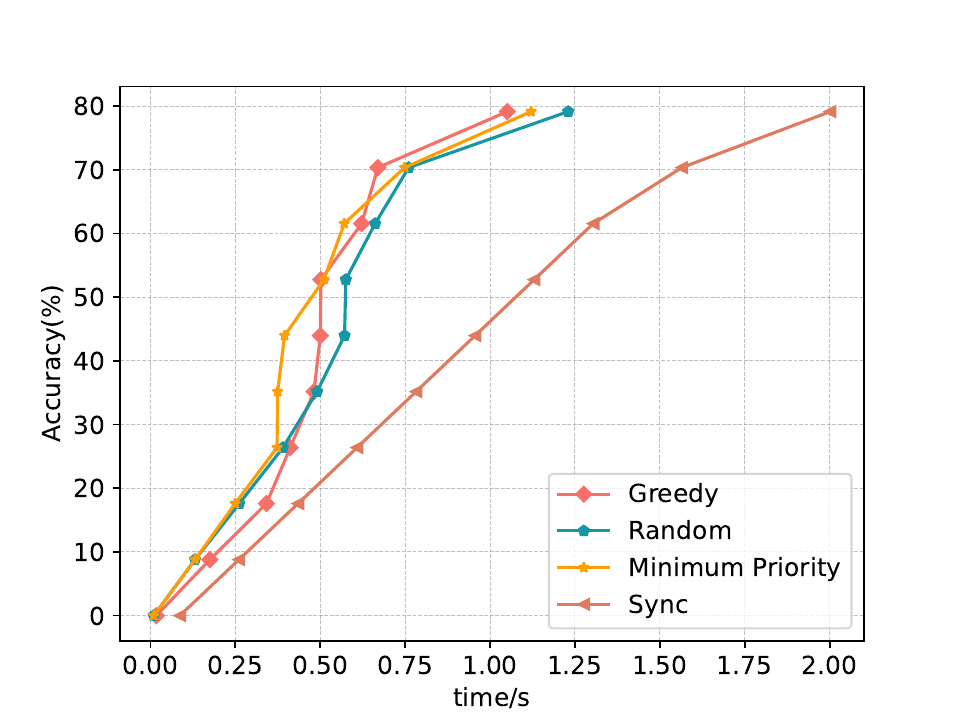}
        \caption{MNIST}
        \label{fig:mnist3}
    \end{subfigure}
    \hfill 
    \begin{subfigure}{0.32\textwidth}
        \includegraphics[width=\linewidth]{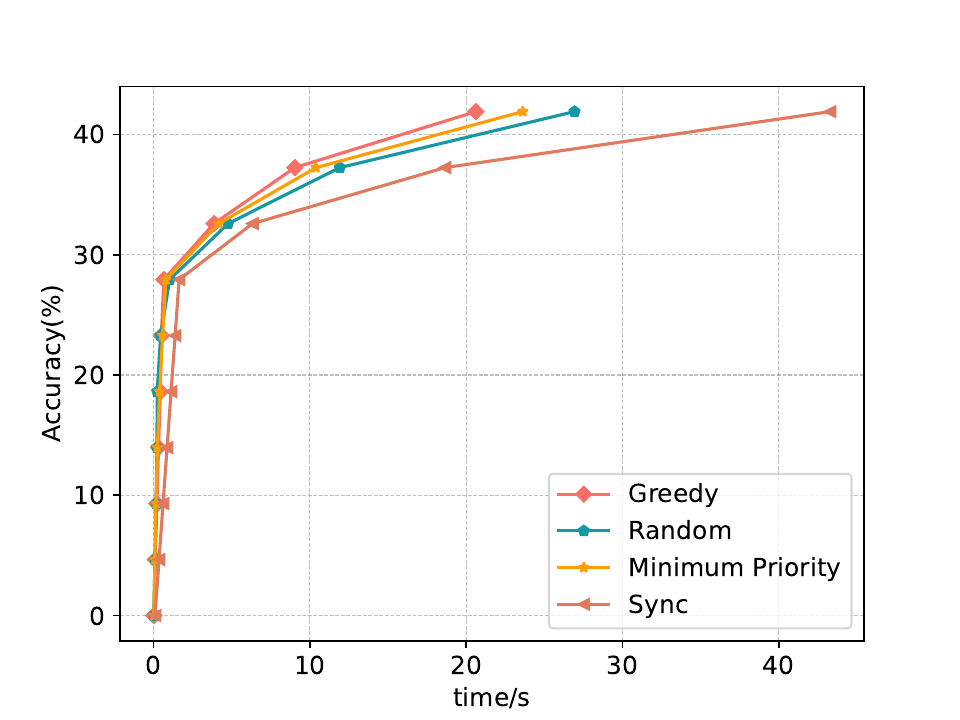}
        \caption{CIFAR-10}
        \label{fig:CIFAR-103}
    \end{subfigure}
    \begin{subfigure}{0.32\textwidth}
        \includegraphics[width=\linewidth]{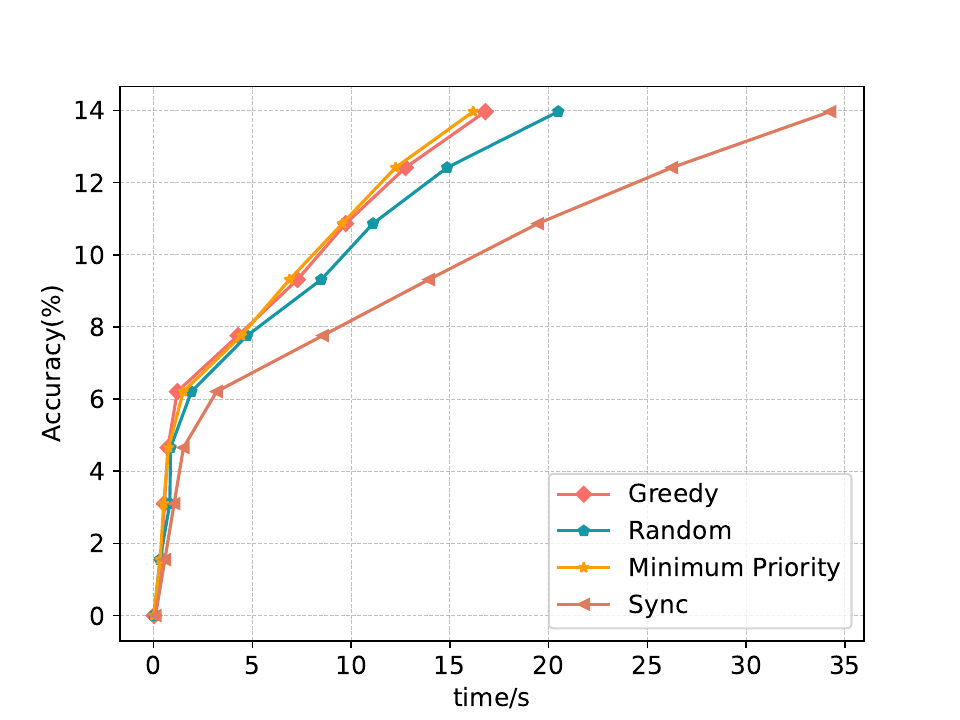}
        \caption{CIFAR-100}
        \label{fig:CIFAR-1003}
    \end{subfigure}
    \hfill 
    \caption{The comparison of ablation methods on different datasets}
    \label{fig:Ablation}
\end{figure}

In addressing the model assignment problem within Fed-RAA, we designed a scheduler-based online algorithm, named Gre-RAA, as outlined in Algorithm~\ref{alg2}. 
Although it has been theoretically affirmed that Gre-RAA achieves the same upper bound of training delay as that achieved by the optimal offline model assignment strategy, it remains imperative to conduct ablation studies to validate the contributions of Gre-RAA toward enhancing both the convergence speed and the maximum accuracy within Fed-RAA. To this end, we instituted two variance algorithms, encompassing: (1) \textit{Random Fed-RAA}: random assignment, and (2) \textit{Minimum Priority Fed-RAA}(MP-RAA): consistently allocating the model segment that has undergone the least amount of training. 
Furthermore, to validate the effectiveness of the asynchronous aggregation approach utilized by Fed-RAA within this framework, we designed a variance algorithm for an ablation study: \textit{Sync Fed-RAA}. This variant employs a synchronous aggregation method for FL, with all other aspects remaining identical to Fed-RAA.
We evaluate the impact of these variance algorithms on Fed-RAA's performance under the conditions of three distinct dataset scenarios. 
The accuracy results from this comparative analysis are depicted in Figure~\ref{fig:Ablation} and Table~\ref{ablation}.
\begin{table}[ht!]
    \caption{Comparison of algorithms on MNIST, CIFAR-10, and CIFAR-100 datasets}
    \label{ablation}
      \centering
\vspace{\baselineskip}
    \begin{tabular}{lcccccc}
    \toprule
     & \multicolumn{2}{c}{MNIST} & \multicolumn{2}{c}{CIFAR-10} & \multicolumn{2}{c}{CIFAR-100} \\
    \cmidrule(r){2-3} \cmidrule(r){4-5} \cmidrule(r){6-7}
    Baseline &  Accuracy (\%) & Time (s) & Accuracy (\%) & Time (s) & Accuracy (\%) & Time (s) \\
    \midrule
        Gre-RAA &  93.95 & 69.46 & 47.43 & 68.58 & 20.74 & 75.96 \\
    Random      &  93.76 & 78.54 & 47.43 & 84.63 & 18.77 & 60.10 \\
    MP-RAA &  93.88 & 80.73 & 47.39 & 84.77 & 20.75 & 78.28 \\
    Sync       &  93.94 & 162.71 & 47.84 & 162.82 & 21.46 & 179.44 \\
    \bottomrule
    \end{tabular}
\end{table}

In our ablation study, we observed distinct performance characteristics among the variants: Random Fed-RAA, Minimum Priority Fed-RAA (MP-RAA), and Sync Fed-RAA. The results demonstrate Fed-RAA's efficiency on the MNIST dataset with a high accuracy of 93.95\% and a low training time of 23.15 seconds. However, on the more complex CIFAR-10 and CIFAR-100 datasets, although Sync Fed-RAA occasionally achieved higher accuracies (47.84\% and 21.46\%, respectively), it consistently demanded significantly more training time, indicating a trade-off between accuracy and computational efficiency. The Random and MP-RAA variants showed competitive performance across all datasets, suggesting their potential utility in scenarios where a balance between accuracy and training time is critical.
This analysis elucidates the nuanced trade-offs involved in different Fed-RAA configurations. While Sync Fed-RAA may offer accuracy advantages, its higher time consumption could limit its applicability in resource-constrained settings. On the other hand, Random and MP-RAA present more balanced solutions, potentially optimizing federated learning outcomes in diverse environments. These insights will guide future developments in federated learning strategies, aiming to adaptively balance accuracy, efficiency, and resource usage across varying conditions.

\subsection{Scalability of Our Approach}
In our experiments, we conducted a detailed investigation into the performance of the Federated Random Assignment Algorithm (Fed-RAA), with particular emphasis on the impact of two critical hyperparameters: the proportion of extremely low-capability users, denoted as \(\beta\), and the total number of sub-models post-segmentation, denoted as \(M\), on the time-accuracy trade-off inherent to the algorithm. 
For ease of description, in this experiment, we categorize the user groups into two based on their capability levels: extremely low-capability users (capability of 1) and low-capability users (capability of 3). The proportion of extremely low-capability users, denoted by \(\beta\), illustrates the heterogeneity in user capabilities; for instance, \(\beta = 0.1\) indicates that 10\% of the users are extremely low-capability, while the remaining 90\% are low-capability users.
\begin{figure}[ht!]
    \centering 
    \begin{subfigure}{0.32\textwidth}
        \includegraphics[width=\linewidth]{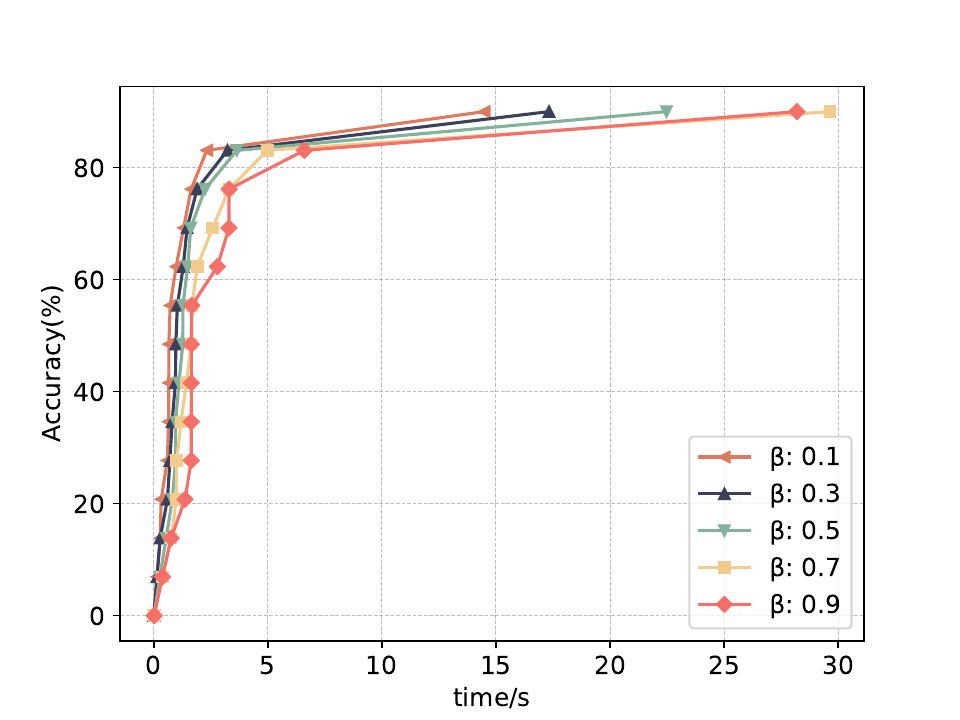}
        \caption{MNIST}
        \label{fig:mnist2}
    \end{subfigure}
    \hfill
    \begin{subfigure}{0.32\textwidth}
        \includegraphics[width=\linewidth]{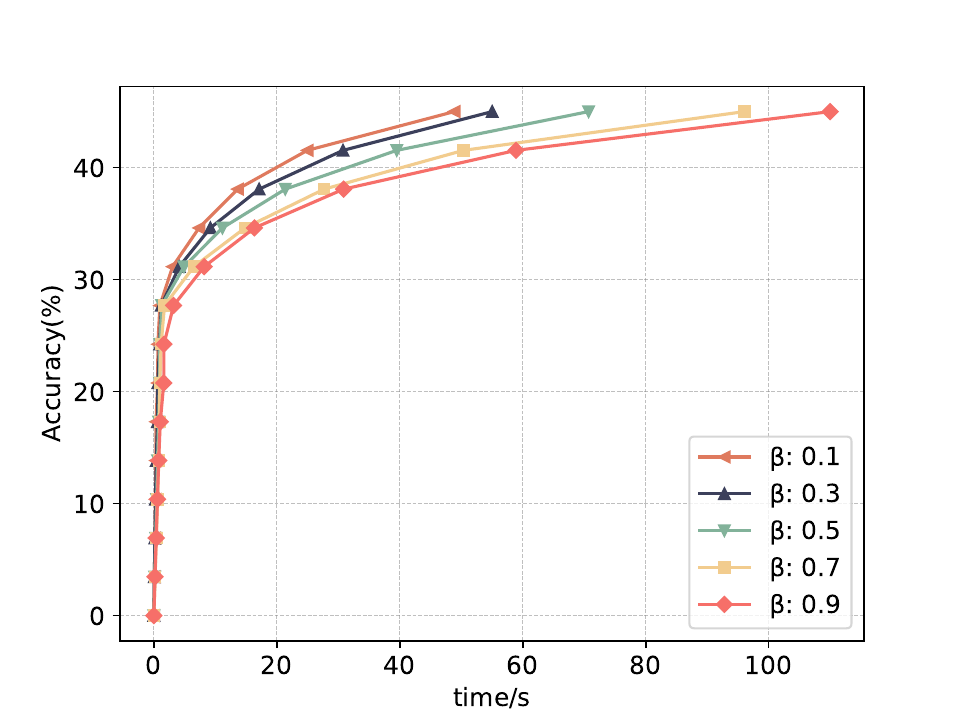}
        \caption{CIFAR-10}
        \label{fig:CIFAR-102}
    \end{subfigure}
    \begin{subfigure}{0.32\textwidth}
        \includegraphics[width=\linewidth]{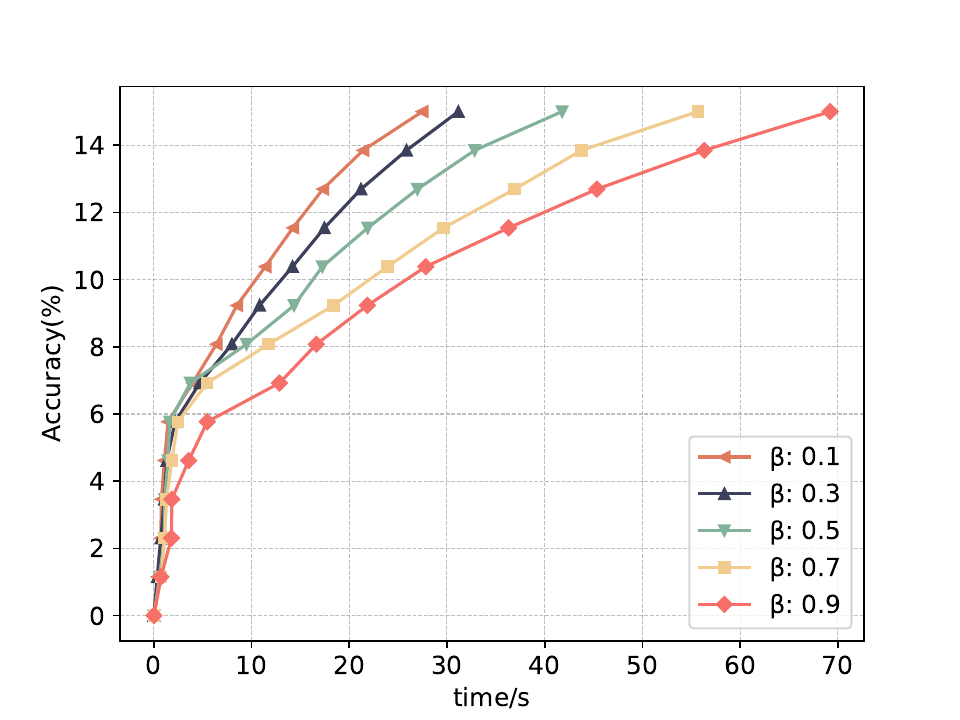}
        \caption{CIFAR-100}
        \label{fig:CIFAR-1002}
    \end{subfigure}
    \hfill
    \caption{The influence of \(\beta\)}
    \label{fig:beta}
\end{figure}
The analysis in Figure~\ref{fig:beta} shows that increasing the proportion of extremely low-capability clients (\(\beta\)) severely affects the training time and accuracy of federated learning models, particularly in complex datasets like CIFAR-100 where training time increases considerably. 
As \(\beta\) increases, training slows significantly due to the higher number of extremely low-capability clients, and accuracy declines more noticeably in complex datasets. 
This emphasizes the substantial impact of client capability distribution on federated learning performance, especially under challenging conditions. 
These findings indicate the necessity for adaptive strategies and careful management of client heterogeneity to improve both efficiency and effectiveness in federated learning across diverse datasets. 

\begin{figure}[ht!]
    \centering 
    \begin{subfigure}{0.32\textwidth}
        \includegraphics[width=\linewidth]{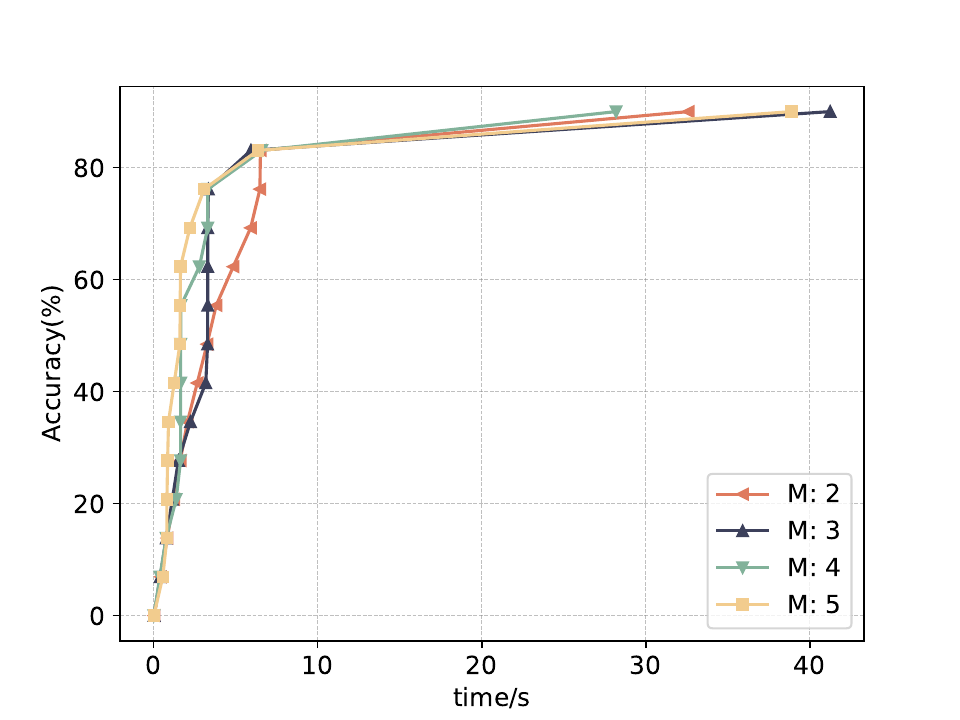}
        \caption{MNIST}
        \label{fig:mnist4}
    \end{subfigure}
    \hfill 
    \begin{subfigure}{0.32\textwidth}
        \includegraphics[width=\linewidth]{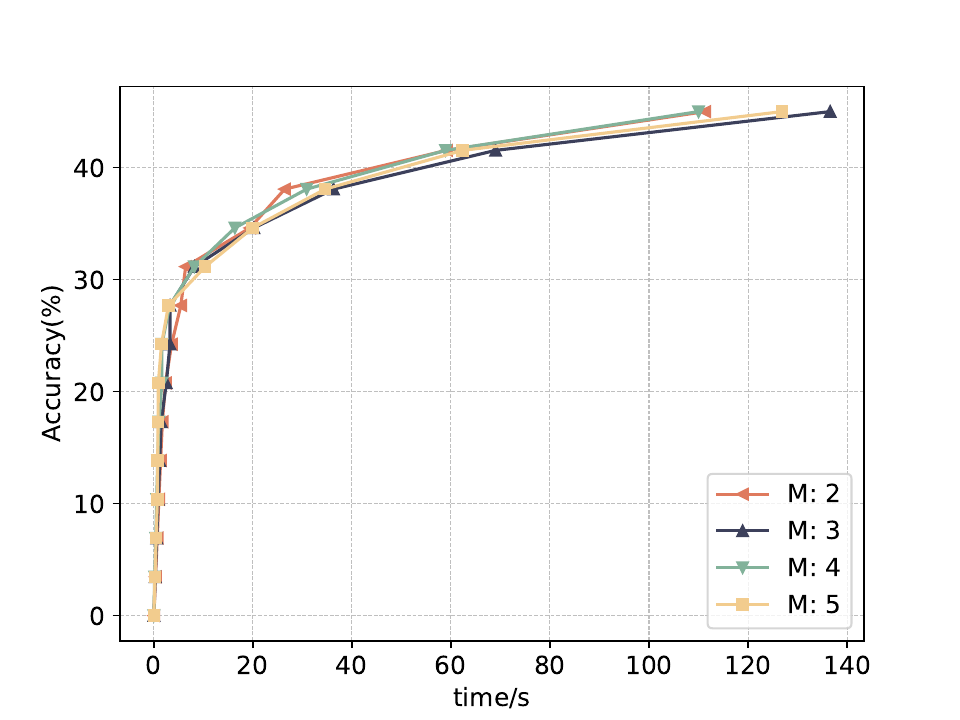}
        \caption{CIFAR-10}
        \label{fig:CIFAR-104}
    \end{subfigure}
    \begin{subfigure}{0.32\textwidth}
        \includegraphics[width=\linewidth]{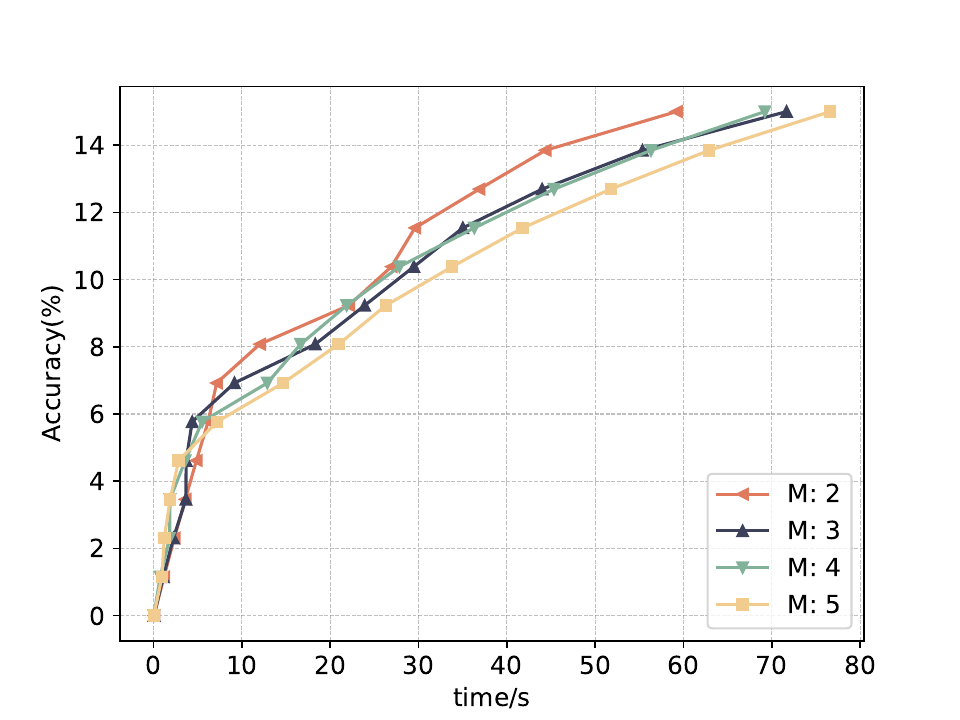}
        \caption{CIFAR-100}
        \label{fig:CIFAR-1004}
    \end{subfigure}
    \hfill 
    \caption{The influence of \(M\)}
    \label{fig:M}
\end{figure}
The results from Figure~\ref{fig:M} highlight the influence of increasing submodel numbers \(M\) on training times, particularly underlining scalability issues in complex datasets like CIFAR-100, where more submodels significantly extend training durations without enhancing accuracy. 
This suggests a diminishing return on training efficiency as the number of submodels increases, especially in complex settings. 
While the MNIST and CIFAR-10 datasets demonstrate only moderate time increases, the accuracy gains do not justify the additional submodels, indicating a crucial trade-off in asynchronous federated learning like Fed-RAA between submodel count and training efficiency. 
These findings emphasize the need to optimize the balance between model fragmentation and overall training efficacy, especially in environments with varied data complexities. 

Moreover, we provide 12 detailed result figures for comparison of \(\beta\) and \(M\), which show similar conclusions as above. 
\section{Conclusion}
This paper studies the federated learning problem among clients with heterogeneous limited resources and proposes a resource-adaptive asynchronous FL algorithm, named Fed-RAA, as the solution. Fed-RAA uniquely assigns model fragments based on individual client capabilities, enabling asynchronous updates and reducing delays from slower clients. This method is validated through robust theoretical analysis and empirical tests on the MNIST, CIFAR-10, and CIFAR-100 datasets, demonstrating substantial gains in learning speed and accuracy. We also employ an online greedy-based algorithm to optimize resource allocation and enhance fairness in local model training. Future work will aim to refine this algorithm and extend its application in federated learning, with a focus on adapting it for specific large model architecture, such as large language model with Mixture of Experts or Lora.
\bibliographystyle{unsrtnat}
\bibliography{ref}

\begin{thebibliography}{28}
\providecommand{\natexlab}[1]{#1}
\providecommand{\url}[1]{\texttt{#1}}
\expandafter\ifx\csname urlstyle\endcsname\relax
  \providecommand{\doi}[1]{doi: #1}\else
  \providecommand{\doi}{doi: \begingroup \urlstyle{rm}\Url}\fi

\bibitem[Artetxe et~al.()Artetxe, Bhosale, Goyal, Mihaylov, Ott, Shleifer, Lin, Du, Iyer, Pasunuru, Anantharaman, Li, Chen, Akin, Baines, Martin, Zhou, Koura, O’Horo, Wang, Zettlemoyer, Diab, Kozareva, and Stoyanov]{Artetxe_Bhosale_Goyal_Mihaylov_Ott_Shleifer_Lin_Du_Iyer_Pasunuru_et}
Mikel Artetxe, Shruti Bhosale, Naman Goyal, Todor Mihaylov, Myle Ott, Sam Shleifer, XiVictoria Lin, Jingfei Du, Srinivasan Iyer, Ramakanth Pasunuru, Giri Anantharaman, Xian Li, Shuohui Chen, Halil Akin, Mandeep Baines, Louis Martin, Xing Zhou, PunitSingh Koura, Brian O’Horo, Jeff Wang, Luke Zettlemoyer, Mona Diab, Zornitsa Kozareva, and Ves Stoyanov.
\newblock Efficient large scale language modeling with mixtures of experts.

\bibitem[Fedus et~al.(2021)Fedus, Zoph, and Shazeer]{Fedus_Zoph_Shazeer_2021}
William Fedus, Barret Zoph, and Noam Shazeer.
\newblock Switch transformers: Scaling to trillion parameter models with simple and efficient sparsity.
\newblock \emph{arXiv: Learning,arXiv: Learning}, Jan 2021.

\bibitem[Lepikhin et~al.(2020)Lepikhin, Lee, Xu, Chen, Firat, Huang, Krikun, Shazeer, and Chen]{Lepikhin_Lee_Xu_Chen_Firat_Huang_Krikun_Shazeer_Chen_2020}
Dmitry Lepikhin, HyoukJoong Lee, Yuanzhong Xu, Dehao Chen, Orhan Firat, Yanping Huang, Maxim Krikun, Noam Shazeer, and Zhifeng Chen.
\newblock Gshard: Scaling giant models with conditional computation and automatic sharding.
\newblock \emph{Cornell University - arXiv,Cornell University - arXiv}, Jun 2020.

\bibitem[Liu and Zhao(2024)]{DBLP:journals/corr/abs-2403-19016}
Chang Liu and Jun Zhao.
\newblock Resource allocation in large language model integrated 6g vehicular networks.
\newblock \emph{CoRR}, abs/2403.19016, 2024.
\newblock \doi{10.48550/ARXIV.2403.19016}.
\newblock URL \url{https://doi.org/10.48550/arXiv.2403.19016}.

\bibitem[Woisetschläger et~al.(2023)Woisetschläger, Isenko, Wang, Mayer, and Jacobsen]{woisetschläger2023federated}
Herbert Woisetschläger, Alexander Isenko, Shiqiang Wang, Ruben Mayer, and Hans-Arno Jacobsen.
\newblock Federated fine-tuning of llms on the very edge: The good, the bad, the ugly, 2023.

\bibitem[McMahan et~al.(2017)McMahan, Moore, Ramage, Hampson, and y~Arcas]{17FL}
Brendan McMahan, Eider Moore, Daniel Ramage, Seth Hampson, and Blaise~Ag{\"{u}}era y~Arcas.
\newblock Communication-efficient learning of deep networks from decentralized data.
\newblock In \emph{{AISTATS} 2017}, volume~54 of \emph{Proceedings of Machine Learning Research}, pages 1273--1282. {PMLR}, 2017.

\bibitem[Nguyen et~al.(2024)Nguyen, Pham, Wong, Nguyen, Nguyen, and Do]{PWNND24}
Quan Nguyen, Hieu~H. Pham, Kok{-}Seng Wong, Phi~Le Nguyen, Truong~Thao Nguyen, and Minh~N. Do.
\newblock Feddct: Federated learning of large convolutional neural networks on resource-constrained devices using divide and collaborative training.
\newblock \emph{{IEEE} Trans. Netw. Serv. Manag.}, 21\penalty0 (1):\penalty0 418--436, 2024.
\newblock \doi{10.1109/TNSM.2023.3314066}.
\newblock URL \url{https://doi.org/10.1109/TNSM.2023.3314066}.

\bibitem[Qin et~al.(2023)Qin, Zhang, Liu, and Qian]{FLL1}
Jiangcheng Qin, Xueyuan Zhang, Baisong Liu, and Jiangbo Qian.
\newblock A split-federated learning and edge-cloud based efficient and privacy-preserving large-scale item recommendation model.
\newblock \emph{J. Cloud Comput.}, 12\penalty0 (1):\penalty0 57, 2023.
\newblock \doi{10.1186/S13677-023-00435-5}.
\newblock URL \url{https://doi.org/10.1186/s13677-023-00435-5}.

\bibitem[Diao et~al.(2020)Diao, Ding, and Tarokh]{Diao_Ding_Tarokh_2020}
Enmao Diao, Jie Ding, and Vahid Tarokh.
\newblock Heterofl: Computation and communication efficient federated learning for heterogeneous clients.
\newblock \emph{arXiv: Learning,arXiv: Learning}, Oct 2020.

\bibitem[Guo et~al.(2021)Guo, Hu, and Hu]{265005}
Peizhen Guo, Bo~Hu, and Wenjun Hu.
\newblock Mistify: Automating {DNN} model porting for {On-Device} inference at the edge.
\newblock In \emph{18th USENIX Symposium on Networked Systems Design and Implementation (NSDI 21)}, pages 705--719. USENIX Association, April 2021.
\newblock ISBN 978-1-939133-21-2.
\newblock URL \url{https://www.usenix.org/conference/nsdi21/presentation/guo}.

\bibitem[Ogden et~al.(2021)Ogden, Gilman, Walls, and Guo]{9659532}
Samuel~S. Ogden, Guin~R. Gilman, Robert~J. Walls, and Tian Guo.
\newblock Many models at the edge: Scaling deep inference via model-level caching.
\newblock In \emph{2021 IEEE International Conference on Autonomic Computing and Self-Organizing Systems (ACSOS)}, pages 51--60, 2021.
\newblock \doi{10.1109/ACSOS52086.2021.00027}.

\bibitem[Thapa et~al.(2022)Thapa, Chamikara, Camtepe, and Sun]{SplitFed}
Chandra Thapa, Mahawaga Arachchige~Pathum Chamikara, Seyit Camtepe, and Lichao Sun.
\newblock Splitfed: When federated learning meets split learning.
\newblock In \emph{Thirty-Sixth {AAAI} Conference on Artificial Intelligence, {AAAI} 2022, Thirty-Fourth Conference on Innovative Applications of Artificial Intelligence, {IAAI} 2022, The Twelveth Symposium on Educational Advances in Artificial Intelligence, {EAAI} 2022 Virtual Event, February 22 - March 1, 2022}, pages 8485--8493. {AAAI} Press, 2022.
\newblock \doi{10.1609/AAAI.V36I8.20825}.
\newblock URL \url{https://doi.org/10.1609/aaai.v36i8.20825}.

\bibitem[Wang et~al.(2023)Wang, Zhang, Li, Lan, Chen, Xiong, Cheng, and Yu]{RA-FED}
Yangyang Wang, Xiao Zhang, Mingyi Li, Tian Lan, Huashan Chen, Hui Xiong, Xiuzhen Cheng, and Dongxiao Yu.
\newblock Theoretical convergence guaranteed resource-adaptive federated learning with mixed heterogeneity.
\newblock In Ambuj~K. Singh, Yizhou Sun, Leman Akoglu, Dimitrios Gunopulos, Xifeng Yan, Ravi Kumar, Fatma Ozcan, and Jieping Ye, editors, \emph{Proceedings of the 29th {ACM} {SIGKDD} Conference on Knowledge Discovery and Data Mining, {KDD} 2023, Long Beach, CA, USA, August 6-10, 2023}, pages 2444--2455. {ACM}, 2023.
\newblock \doi{10.1145/3580305.3599521}.
\newblock URL \url{https://doi.org/10.1145/3580305.3599521}.

\bibitem[Xu et~al.(2023)Xu, Wu, Cai, Li, and Wang]{fine-tuning-LLM2023}
Mengwei Xu, Yaozong Wu, Dongqi Cai, Xiang Li, and Shangguang Wang.
\newblock Federated fine-tuning of billion-sized language models across mobile devices.
\newblock \emph{CoRR}, abs/2308.13894, 2023.
\newblock \doi{10.48550/ARXIV.2308.13894}.
\newblock URL \url{https://doi.org/10.48550/arXiv.2308.13894}.

\bibitem[Yuan et~al.(2022)Yuan, Wolfe, Dun, Tang, Kyrillidis, and Jermaine]{Yuan_Wolfe_Dun_Tang_Kyrillidis_Jermaine_2022}
Binhang Yuan, Cameron~R. Wolfe, Chen Dun, Yuxin Tang, Anastasios Kyrillidis, and Chris Jermaine.
\newblock Distributed learning of fully connected neural networks using independent subnet training.
\newblock \emph{Proceedings of the VLDB Endowment}, page 1581–1590, Apr 2022.
\newblock \doi{10.14778/3529337.3529343}.
\newblock URL \url{http://dx.doi.org/10.14778/3529337.3529343}.

\bibitem[Yang et~al.(2023)Yang, Xiao, Motta, Beaufays, Mathews, and Chen]{FLL2}
Tien{-}Ju Yang, Yonghui Xiao, Giovanni Motta, Fran{\c{c}}oise Beaufays, Rajiv Mathews, and Mingqing Chen.
\newblock Online model compression for federated learning with large models.
\newblock In \emph{{IEEE} International Conference on Acoustics, Speech and Signal Processing {ICASSP} 2023, Rhodes Island, Greece, June 4-10, 2023}, pages 1--5. {IEEE}, 2023.
\newblock \doi{10.1109/ICASSP49357.2023.10097124}.
\newblock URL \url{https://doi.org/10.1109/ICASSP49357.2023.10097124}.

\bibitem[Jiang et~al.(2022{\natexlab{a}})Jiang, Wang, Valls, Ko, Lee, Leung, and Tassiulas]{pruning_2022}
Yuang Jiang, Shiqiang Wang, Victor Valls, Bong~Jun Ko, Wei-Han Lee, Kin~K. Leung, and Leandros Tassiulas.
\newblock Model pruning enables efficient federated learning on edge devices.
\newblock \emph{IEEE Transactions on Neural Networks and Learning Systems}, page 1–13, Jan 2022{\natexlab{a}}.
\newblock \doi{10.1109/tnnls.2022.3166101}.
\newblock URL \url{http://dx.doi.org/10.1109/tnnls.2022.3166101}.

\bibitem[Lin et~al.(2017)Lin, Han, Mao, Wang, and Dally]{lin2017deep}
Yujun Lin, Song Han, Huizi Mao, Yu~Wang, and William~J Dally.
\newblock Deep gradient compression: Reducing the communication bandwidth for distributed training.
\newblock \emph{arXiv preprint arXiv:1712.01887}, 2017.

\bibitem[Jiang et~al.(2022{\natexlab{b}})Jiang, Wang, Valls, Ko, Lee, Leung, and Tassiulas]{fedprun}
Yuang Jiang, Shiqiang Wang, Victor Valls, Bong~Jun Ko, Wei-Han Lee, Kin~K Leung, and Leandros Tassiulas.
\newblock Model pruning enables efficient federated learning on edge devices.
\newblock \emph{IEEE Transactions on Neural Networks and Learning Systems}, 2022{\natexlab{b}}.

\bibitem[Krizhevsky et~al.(2009)Krizhevsky, Hinton, et~al.]{CIFAR}
Alex Krizhevsky, Geoffrey Hinton, et~al.
\newblock Learning multiple layers of features from tiny images.
\newblock 2009.

\bibitem[Deng(2012)]{MNIST}
Li~Deng.
\newblock The mnist database of handwritten digit images for machine learning research.
\newblock \emph{IEEE Signal Processing Magazine}, 29\penalty0 (6):\penalty0 141--142, 2012.

\bibitem[McMahan et~al.(2016)McMahan, Moore, Ramage, and y~Arcas]{FedAvg}
H.~Brendan McMahan, Eider Moore, Daniel Ramage, and Blaise~Ag{\"{u}}era y~Arcas.
\newblock Federated learning of deep networks using model averaging.
\newblock \emph{CoRR}, abs/1602.05629, 2016.
\newblock URL \url{http://arxiv.org/abs/1602.05629}.

\bibitem[Xie et~al.(2019)Xie, Koyejo, and Gupta]{fedsync}
Cong Xie, Sanmi Koyejo, and Indranil Gupta.
\newblock Asynchronous federated optimization.
\newblock \emph{arXiv preprint arXiv:1903.03934}, 2019.

\bibitem[Li et~al.(2020)Li, Sahu, Zaheer, Sanjabi, Talwalkar, and Smith]{fedprox}
Tian Li, Anit~Kumar Sahu, Manzil Zaheer, Maziar Sanjabi, Ameet Talwalkar, and Virginia Smith.
\newblock Federated optimization in heterogeneous networks.
\newblock \emph{Proceedings of Machine learning and systems}, 2:\penalty0 429--450, 2020.

\bibitem[Zhang et~al.(2024{\natexlab{a}})Zhang, Zhou, Imani, Lee, and Lan]{zhang2024collaborative}
Zuyuan Zhang, Hanhan Zhou, Mahdi Imani, Taeyoung Lee, and Tian Lan.
\newblock Collaborative ai teaming in unknown environments via active goal deduction.
\newblock \emph{arXiv preprint arXiv:2403.15341}, 2024{\natexlab{a}}.

\bibitem[Zou et~al.(2024)Zou, Zhang, Zhang, Zheng, Yu, and Yu]{zou2024distributed}
Yifei Zou, Zuyuan Zhang, Congwei Zhang, Yanwei Zheng, Dongxiao Yu, and Jiguo Yu.
\newblock A distributed abstract mac layer for cooperative learning on internet of vehicles.
\newblock \emph{IEEE Transactions on Intelligent Transportation Systems}, 2024.

\bibitem[Zhang et~al.(2024{\natexlab{b}})Zhang, Imani, and Lan]{zhang2024modeling}
Zuyuan Zhang, Mahdi Imani, and Tian Lan.
\newblock Modeling other players with bayesian beliefs for games with incomplete information.
\newblock \emph{arXiv preprint arXiv:2405.14122}, 2024{\natexlab{b}}.

\bibitem[Gao et~al.(2024)Gao, Zou, Zhang, Cheng, and Yu]{gao2024cooperative}
Mengtong Gao, Yifei Zou, Zuyuan Zhang, Xiuzhen Cheng, and Dongxiao Yu.
\newblock Cooperative backdoor attack in decentralized reinforcement learning with theoretical guarantee.
\newblock \emph{arXiv preprint arXiv:2405.15245}, 2024.

\end{thebibliography}

%%%%%%%%%%%%%%%%%%%%%%%%%%%%%%%%%%%%%%%%%%%%%%%%%%%%%%%%%%%%
\newpage

\appendix

% \section{Appendix}
\section{Convergence Analysis of Fed-RAA}
Let us start the convergence proof of Fed-RAA: 

    Without loss of generality, we assume that in the \(q^{th}\) epoch of model \(\theta^j\) for each \(j\in [M]\), the server receives the model \(\theta^j_{new}\), with the time stamp \(\tau\). 
    We assume that \(\theta^j_{new}\) is the result of applying \(I_{min}^n\leq I^n\leq I_{max}^n\) local updates to \(\theta_{\tau}^j\) on the \(n\)th device. 
    For convenience, we ignore \(n\) in all variances above and also in \(z^n_{\tau,i}\). 
    Thus, using smoothness and strong convexity, conditional on \(\theta_{\tau,i-1}\), for \(\forall i\in [I]\) we have 
    \begin{align*}
    \mathbb{E}[F(\theta^j_{\tau,i})-F(\theta_{*})] &\leq \mathbb{E}[G_{\theta^j_{\tau}}(\theta^j_{\tau,i})-F(\theta_{*})] \\
    &\leq G_{\theta^j_{\tau}}(\theta^j_{\tau,i-1})-F(\theta_{*})-\gamma\mathbb{E}[\langle\nabla G_{\theta^j_{\tau}}(\theta^j_{\tau,i-1}),\nabla g_{\theta^j_{\tau}}(\theta^j_{\tau,i-1};z_{\tau,i})  \rangle]\\
    &\ \ \ \ +\frac{L\gamma^2}{2}\mathbb{E}[\|g_{\theta^j_{\tau}}(\theta^j_{\tau,i-1};z_{\tau,i})\|^2] \\
    &\leq F(\theta^j_{\tau,i-1})-F(\theta_{*}) +\frac{\rho}{2}\|\theta^j_{\tau,i-1}-\theta^j_\tau\|^2\\
    &\ \ \ \ -\gamma\mathbb{E}[\langle\nabla G_{\theta^j_{\tau}}(\theta^j_{\tau,i-1}),\nabla g_{\theta^j_{\tau}}(\theta^j_{\tau,i-1};z_{\tau,i})  \rangle]+\frac{L\gamma^2}{2}\mathbb{E}[\|g_{\theta^j_{\tau}}(\theta^j_{\tau,i-1};z_{\tau,i})\|^2] \\
    &\leq F(\theta^j_{\tau,i-1})-F(\theta_{*})-\gamma\mathbb{E}[\langle\nabla G_{\theta^j_{\tau}}(\theta^j_{\tau,i-1}),\nabla g_{\theta^j_{\tau}}(\theta^j_{\tau,i-1};z_{\tau,i})  \rangle]\\
    &\ \ \ \ +\frac{L\gamma^2}{2}V_2 + \frac{\rho I_{max}^2\gamma^2}{2}V_2 \\
    &\leq F(\theta^j_{\tau,i-1})-F(\theta_{*})-\gamma\mathbb{E}[\langle\nabla G_{\theta^j_{\tau}}(\theta^j_{\tau,i-1}),\nabla g_{\theta^j_{\tau}}(\theta^j_{\tau,i-1};z_{\tau,i})  \rangle]\\
    &\ \ \ \ + \gamma^2\mathcal{O}(\rho I_{max}^2V_2)
    \end{align*}

    Taking \(\rho\) large enough such that \(-(1+2\rho+\epsilon)V_1+\rho^2\|\theta^j_{\tau,i}-\theta^j_{\tau}\|^2-\frac{\rho}{2}\|\theta^j_{\tau,i}-\theta^j_{\tau}\|^2\geq 0, \forall \theta^j_{\tau,i},\theta^j_{\tau}\), and write \(\nabla g_{\theta^j_{\tau}}(\theta^j_{\tau,i-1};z_{\tau,i})\) as \(\nabla g_{\theta^j_{\tau}}(\theta^j_{\tau,i-1})\) for convenience, we have 

    \begin{align*}
    \langle\nabla G_{\theta^j_{\tau}}(\theta^j_{\tau,i-1}),\nabla g_{\theta^j_{\tau}}(\theta^j_{\tau,i-1})\rangle& -\epsilon\|\nabla F(\theta^j_{\tau,i-1})\|^2 \\
    &= \langle\nabla F(\theta^j_{\tau,i-1})+\rho(\theta^j_{\tau,i-1}-\theta^j_\tau),\nabla f(\theta^j_{\tau,i-1})+\rho(\theta^j_{\tau,i-1}-\theta^j_\tau) \rangle \\
    &\quad -\epsilon\|\nabla F(\theta^j_{\tau,i-1})\|^2 \\
    &= \langle\nabla F(\theta^j_{\tau,i-1}),\nabla f(\theta^j_{\tau,i-1})\rangle  \\
    &\quad + \rho\langle\nabla F(\theta^j_{\tau,i-1})+\nabla f(\theta^j_{\tau,i-1}),(\theta^j_{\tau,i-1}-\theta^j_\tau)\rangle\\
    &\quad + \rho^2\|\theta^j_{\tau,i-1}-\theta^j_\tau\|^2 -\epsilon\|\nabla F(\theta^j_{\tau,i-1})\|^2 \\
    &\geq -\frac{1}{2}\|\nabla F(\theta^j_{\tau,i-1})\|^2 -\frac{1}{2}\|\nabla f(\theta^j_{\tau,i-1})\|^2\\
    &\quad -\frac{\rho}{2}\|\nabla F(\theta^j_{\tau,i-1})+\nabla f(\theta^j_{\tau,i-1})\|^2 \\
    &\quad -\frac{\rho}{2}\|\theta^j_{\tau,i-1}-\theta^j_\tau\|^2 + \rho^2\|\theta^j_{\tau,i-1}-\theta^j_\tau\|^2 -\epsilon\|\nabla F(\theta^j_{\tau,i-1})\|^2 \\
    &\geq -(1+2\rho+\epsilon)V_1 + \rho^2\|\theta^j_{\tau,i}-\theta^j_{\tau}\|^2 -\frac{\rho}{2}\|\theta^j_{\tau,i}-\theta^j_{\tau}\|^2 \\
    &= a\rho^2 + b\rho + c \geq 0,
    \end{align*}
    
    where \(a=\|\theta^j_{\tau,i}-\theta^j_{\tau}\|^2>0\), \(b=-2V_1-\frac{1}{2}\|\theta^j_{\tau,i}-\theta^j_{\tau}\|^2\), \(c=-(1+\epsilon)V_1\). Thus, we have \(-\gamma\langle\nabla G_{\theta^j_{\tau}}(\theta^j_{\tau,i-1}),\nabla g_{\theta^j_{\tau}}(\theta^j_{\tau,i-1})\rangle\leq -\gamma \epsilon\|\nabla F(\theta^j_{\tau,i-1})\|^2\).

    Using \(\tau-(q-1)\leq K\), we have \(\|\theta_\tau-\theta_{q-1}\|^2\)\(\leq\|(\theta_\tau-\theta_{\tau+1})+\ldots+(\theta_{q-1}-\theta_{q-1})\|^2\leq K\|\theta_\tau-\theta_{\tau+1}\|^2 + \ldots+K\|\theta_{q-1}-\theta_{q-1}\|^2\)\(\leq\alpha^2\gamma^2K^2I_{max}^2\mathcal{O}(V_2)\).

    Also, we have \(\|\theta_\tau-\theta_{q-1}\|\)\(\leq\|(\theta_\tau-\theta_{\tau+1})+\ldots+(\theta_{q-1}-\theta_{q-1})\|\leq \|\theta_\tau-\theta_{\tau+1}\|^2 + \ldots+\|\theta_{q-1}-\theta_{q-1}\|^2\)\(\leq\alpha\gamma KI_{max}\mathcal{O}(\sqrt{V_2})\). 

    Thus, we have 
    \begin{align*}
    \mathbb{E}[F(\theta^j_{\tau,i})-F(\theta_{*})] &\\
    \leq F(\theta^j_{\tau,i-1})-F(\theta_{*}) &-\gamma\mathbb{E}[\langle\nabla G_{\theta^j_{\tau}}(\theta^j_{\tau,i-1}),\nabla g_{\theta^j_{\tau}}(\theta^j_{\tau,i-1};z_{\tau,i})  \rangle] + \gamma^2\mathcal{O}(\rho I_{max}^2V^2) \\
    \leq F(\theta^j_{\tau,i-1})-F(\theta_{*}) &-\gamma \epsilon\|\nabla F(\theta^j_{\tau,i-1})\|^2 + \gamma^2\mathcal{O}(\rho I_{max}^2V_2)
    \end{align*}
    where \(\theta^j_{\tau}\) can be also written as \(\theta_{\tau}\).
    
    By rearranging the terms and telescoping, we have 
    \begin{align*}
    \mathbb{E}[F(\theta^j_{\tau,I})-F(\theta_{\tau})] &\leq -\gamma\epsilon\sum_{i=0}^{I-1}{\mathbb{E}\|\nabla F(\theta^j_{\tau,i})\|^2} + \gamma^2\mathcal{O}(\rho I_{max}^3V_2).
    \end{align*}

    Then, we have     
    \begin{align*}
    \mathbb{E}[F(\theta_{q})-F(\theta_{q-1})] &\leq \mathbb{E}[G_{\theta_{q-1}}(\theta_{q})-F(\theta_{q-1})] \\
    &\leq \mathbb{E}[(1-\alpha)G_{\theta_{q-1}}(\theta_{q-1})+\alpha G_{\theta_{q-1}}(\theta^j_{\tau,I})-F(\theta_{q-1})] \\
    &\leq \mathbb{E}[\alpha(F(\theta^j_{\tau,I})-F(\theta_{q-1}))+\frac{\alpha\rho}{2}\|\theta^j_{\tau,I}-\theta_{q-1}\|^2] \\
    &\leq \alpha\mathbb{E}[F(\theta^j_{\tau,I})-F(\theta_{q-1})]+\alpha\rho\|\theta^j_{\tau,I}-\theta_\tau\|^2 + \alpha\rho\|\theta_{\tau}-\theta_{q-1}\|^2 \\
    &\leq \alpha\mathbb{E}[F(\theta^j_{\tau,I})-F(\theta_{q-1})]+\alpha\rho(\gamma^2I_{max}^2\mathcal{O}(V_2)+\alpha^2\gamma^2K^2I_{max}^2\mathcal{O}(V_2)) \\
    &\leq \alpha\mathbb{E}[F(\theta^j_{\tau,I})-F(\theta_{q-1})]+\alpha\rho(\gamma^2K^2I_{max}^2\mathcal{O}(V_2)) \\
    &\leq \alpha\mathbb{E}[F(\theta^j_{\tau,I})-F(\theta_\tau)+F(\theta_\tau)-F(\theta_{q-1})]+\alpha\gamma^2K^2I_{max}^2\mathcal{O}(V_2)
    \end{align*}

    Using \(L\)-smoothness, we have 
    \begin{align*}
    F(\theta_\tau)-F(\theta_{q-1}) &\leq \langle\nabla F(\theta_{q-1}),\theta_{\tau}-\theta_{q-1}\rangle + \frac{L}{2}\|\theta_{\tau}-\theta_{q-1}\|^2 \\
    &\leq \|\nabla F(\theta_{q-1})\|\|\theta_{\tau}-\theta_{q-1}\|+\frac{L}{2}\|\theta_{\tau}-\theta_{q-1}\|^2 \\
    &\leq \sqrt{V_1}\alpha\gamma KI_{max}\mathcal{O}(\sqrt{V_2})+\frac{L}{2}\alpha^2\gamma^2K^2I_{max}^2\mathcal{O}(V_2) \\
    &\leq \alpha\gamma KI_{max}\mathcal{O}(\sqrt{V_1V_2}) + \alpha^2\gamma^2K^2I_{max}^2\mathcal{O}(V_2)
    \end{align*}
    
    Thus, we have 
    \begin{align*}
    \mathbb{E}[F(\theta_q)-F(\theta_{q-1})] \leq & -\alpha\gamma\epsilon\sum_{i=1}^{I-1}{\mathbb{E}\|\nabla F(\theta^j_{\tau,i})\|^2} \\
    & +\alpha\gamma^2\mathcal{O}(\rho I_{max}^3V_2) \\
    & +\alpha^2\gamma KI_{max}\mathcal{O}(\sqrt{V_1V_2}) \\
    & + \alpha^3\gamma^2K^2I_{max}^2\mathcal{O}(V_2) \\
    & +\alpha\gamma^2K^2I_{max}^2\mathcal{O}(V_2)
    \end{align*}

    By rearranging the terms, we have 
    \begin{align*}
    \sum_{i=1}^{I_q'-1}{\mathbb{E}\|\nabla F(\theta^j_{\tau,i})\|^2} \leq & \frac{\mathbb{E}[F(\theta_{q-1})-F(\theta_{q})]}{\alpha\gamma\epsilon} \\
    & + \frac{\gamma I_{max}^3}{\epsilon}\mathcal{O}(V_2) \\
    & +\frac{\alpha K I_{max}}{\epsilon}\mathcal{O}(\sqrt{V_1V_2}) \\
    & +\frac{\alpha^2\gamma K^2I_{max}^2}{\epsilon}\mathcal{O}(V_2) \\
    & + \frac{\gamma K^2 I_{max}^2}{\epsilon}\mathcal{O}(V_2),
    \end{align*}

    where \(I_q'\) is the number of local iterations applied in the \(q\)th iteration. 

    By telescoping and taking total expectation, after \(T=Q*M\) global epochs which means \(Q\) global epochs on each submodel \(\theta^j\), we have
    \begin{align*}
    \min_{q=0}^{T-1}\mathbb{E}[\|\nabla F(\theta_q)\|^2] & \leq \frac{1}{\sum_{q=1}^{T}{I_{q}'}}\sum_{q=1}^{T}\sum_{i=0}^{I_q'-1}\|\nabla F(\theta_{\tau,i})\|^2 \\
    & \leq \frac{\mathbb{E}[F(\theta_0)-F(\theta_{T})]}{\alpha\gamma\epsilon TI_{min}} 
        +\frac{\gamma TI_{max}^3}{\epsilon TI_{min}}\mathcal{O}(V_2)
        +\frac{\alpha KTI_{max}}{\epsilon TI_{min}}\mathcal{O}(\sqrt{V_1V_2})\\
    &\quad +\frac{\alpha^2\gamma K^2TI_{max}^2}{\epsilon TI_{min}}\mathcal{O}(V_2)
        +\frac{\gamma K^2TI_{max}^2}{\epsilon TI_{min}}\mathcal{O}(V_2) \\
    & \leq \frac{\mathbb{E}[F(\theta_0)-F(\theta_T)]}{\alpha\gamma\epsilon TI_{min}} 
        +\mathcal{O}(\frac{\gamma I_{max}^3}{\epsilon I_{min}})
        +\mathcal{O}(\frac{\alpha KI_{max}}{\epsilon I_{min}})
        +\mathcal{O}(\frac{\alpha^2\gamma K^2I_{max}^2}{\epsilon I_{min}})\\
    &\quad
        +\mathcal{O}(\frac{\gamma K^2I_{max}^2}{\epsilon I_{min}})
    \end{align*}

    Using \(\delta=\frac{I_{max}}{I_{min}}\), and taking \(\alpha = \frac{1}{\sqrt{I_{min}}}\), \(\gamma = \frac{1}{\sqrt{T}}\), \(T = I_{min}^5\), we have 
    \begin{align*}
    \min_{q=0}^{T-1}\mathbb{E}[\|\nabla F(\theta_q)\|^2] & \leq \mathcal{O}\left(\frac{1}{\epsilon I_{min}^3}\right) + \mathcal{O}\left(\frac{\delta^3}{\epsilon\sqrt{I_{min}}}\right) + \mathcal{O}\left(\frac{K\delta}{\epsilon\sqrt{I_{min}}}\right) \\
    & \quad + \mathcal{O}\left(\frac{K^2\delta^2}{\epsilon\sqrt{I_{min}^5}}\right) + \mathcal{O}\left(\frac{K^2\delta^2}{\epsilon\sqrt{I_{min}^3}}\right).
    \end{align*}

%%%%%%%%%%%%%%%%%%%%%%%%%%%%%%%%%%%%%%%%%%%%%%%%%%%%%%%%%%%%
\section{Details of Gre-RAA}
\begin{algorithm}[htbp]

\caption{Gre-RAA} 
 \label{alg2}
	\KwIn{\(n\) clients, model \(\theta\) with \(M\) regions \(\{\theta^1,\ldots,\theta^M\}\), training delay bound \(K\), global epoch \(t\in [T]\)} 
    \KwOut{\(h_t(n)\)}
        \SetAlgoLined
        \For{each idle client \(n\in [N]\) in \(t\)}{
        Initialize \(c_{n,j}\), \(\forall j\in [M]\) \;
        Find one submodel set \(S = \{\theta^j\|c_{n,j}\leq K\}\)\;
        Set \(h_t(n)\gets j\) where \(\theta^j\) is randomly chosen from \(S\)\;
        \For{each submodel \(\theta^{j'}\in S\)}{
        \(q(j')\) is the number of times \(\theta^{j'}\) has been globally updated\;
        \If{\(q(j')<q(h_t(n))\)}{
        \(h_t(n)\gets j'\)
        }
        }
        }
\end{algorithm}

\textbf{Proof of Theorem~\ref{opti}. }
This part shows the detailed proof of Theorem~\ref{opti}: 

    Let \(K'\) be the upper bound of training delay which follows an offline submodel assignment strategy in the same environment. 
    The whole proof process is as follows: 
    We figure out what \(K\) is like following an offline model assignment strategy in our proposed FL system. 
    We then show that given \(K\), the online model assignment algorithm, {\tt Algorithm}~\ref{alg2}, always has a feasible solution. 

    We first introduce the offline model assignment strategy. 
    Given \(N\) clients and \(M\) regions, the offline model assignment strategy assigns the submodel with a fixed index to each user. 
    If \(N < M\), we can simply apply the idea of time division multiplexing for the strategy where each client is cyclically assigned to submodels within a fixed set. Then \(K\) can be simply computed in the same way as that in situation \(N\geq M\). 
    So here we consider the situation where \(N\geq M\). 
    To begin with, we rank \(N\) clients according to their computational resources and get a sorted client list where \(C_i\geq C_{i+1}, \forall i\in [N-1]\) and \(C_i\) is the computational ability of client \(i\). 
    Similarly, we sort the submodels according to the model size and get a submodel list where \(\|\theta^j\|\geq \|\theta^{j+1}\|, \forall j\in [M-1]\). 
    In the offline model assignment strategy, for each submodel \(\theta^j\), we assign it to the client set \(\{(j-1)*\lfloor \frac{N}{M} \rfloor+1,\ldots,\min{(j*\lfloor \frac{N}{M} \rfloor,N)}\}\). 
    Then the training delay bound \(K\) can be computed as \(K=\max{(\frac{\theta^1}{C_1},\frac{\theta^1}{C_1},\ldots,\frac{\theta^1}{C_{\lfloor \frac{N}{M} \rfloor}},\ldots,\frac{\theta^M}{C_N})}\). 

    We can see that there would not exist one \(K'\) that is smaller than \(K\). 
    If we swap any two elements in the sorted client list, the new \(K'\) would be \(\max(K,\frac{\theta^j}{C_{i'}},\frac{\theta^{j'}}{C_i})\) and obviously \(K'\geq K\). 

    Then, we show that given \(K\) which is figured out in the offline scenario, the proposed online model assignment algorithm always has a feasible solution. 
    That is, {\tt Algorithm}~\ref{alg2} always can find one non-empty submodel set \(S\). 
    It is intuitive that there always exists element \(\theta^{\lceil\frac{nM}{N}\rceil}\) in the set \(S\) when considering idle client \(n\). 
    Therefore, it always can find one feasible submodel assigned to each client. 

    So the proposed model assignment algorithm, shown in {\tt Algorithm}~\ref{alg2}, reaches the same upper bound of training delay as that conducted by the optimal offline model assignment strategy. 

\section{Experiment Details}
\subsection{Experimental Setup}
\textbf{Dataset. }
The CIFAR-10~\cite{CIFAR} dataset consists of 60,000 32x32 color images spanning 10 classes, with 50,000 for training and 10,000 for testing, ensuring equal representation across classes. 
CIFAR-100~\cite{CIFAR} expands upon CIFAR-10 with the same image resolution but includes 100 classes categorized into 20 superclasses, each class providing 500 training and 100 test images labeled with both class and superclass identifiers. 
MNIST~\cite{MNIST} remains a cornerstone in the field, featuring 70,000 normalized 28x28 grayscale images of handwritten digits, with a training set of 60,000 and a test set of 10,000, sourced from a diverse population for reliable model evaluation. 

\textbf{Hardware and Software. }We implement the algorithmic simulation via PyTorch employing multi-threading, wherein each thread sustains a local client model in shared memory for individual training, while the master program oversees the global model. During the algorithm's initialization phase, the simulation introduces random latency based on clients' computational capabilities to emulate the feedback delay of slow devices. The simulation software operates on a Linux server equipped with an AMD EPYC 9654 96-Core Processor clocked at 3.463GHz, an NVIDIA RTX 4090 GPU with 24GB VRAM, and 60GB of system memory.

\textbf{Models. }
Taking into account the varying complexity of datasets, our simulation experiments evaluate the performance of both MLP and CNN architectures as global models. For MLP, the number of hidden neurons is configured as 200 for MNIST, 2048 for CIFAR-10, and 4096 for CIFAR-100. 

\textbf{Model Splitting Strategy. }
In the Fed-RAA framework, sub-models are formed by a subset of neurons taken from the global model, enabling clients to train a reduced set of parameters to enhance the convergence speed. This is realized by implementing a partitioning strategy on the global model, which is governed by predefined hyperparameters and results in the model being divided into distinct segments. Each client, guided by its computational performance, selects an appropriate sub-model for its training endeavors. We have devised four distinct partition strategies, where the global model's designated layers are segmented into 2, 3, 4, or 5 parts respectively, with the detailed allocation of these partitions outlined in Table \ref{tab:model_partition_strategy}. To ensure compatibility across a wide range of client capabilities, the partitioning strategy employs a non-uniform approach: weaker devices are assigned smaller sub-models to minimize total training time, whereas more capable clients receive larger sub-models to optimally leverage their superior processing power.

\begin{table}[h]
    \caption{model partition strategy}
    \label{tab:model_partition_strategy}
    \centering
    \vspace{\baselineskip}
    \begin{tabular}{cl}
        \toprule
        Number of partitions & Partition ratio \\
        \midrule
        2 & 40\%, 60\% \\
        3 & 20\%, 30\%, 50\% \\
        4 & 10\%, 20\%, 30\%, 40\% \\
        5 & 5\% , 10\%, 20\%, 30\%, 35\% \\
        \bottomrule
    \end{tabular}

\end{table}
\textbf{Other Settings. }
In the controlled experiments comparing Fed-RAA with these baselines, we employ the momentum SGD optimizer for training across 10 clients, utilizing a batch size of 128 for MNIST and CIFAR-10 datasets, and 64 for CIFAR-100, with a momentum parameter set to 0.5. Each client is tasked with 5 local epochs per global iteration, and the learning rate is fixed at 0.005. To ensure a fair comparison, identical data partitioning schemes, and hardware configurations are maintained across all methodologies.
\subsection{Baselines Comparison Analysis}
Upon meticulous scrutiny of the experimental data encapsulated in Tables~\ref{comparison_across_datasets} and~\ref{time_epoch_comparison}, our Fed-RAA algorithm emerges as conspicuously advantageous in terms of time efficiency relative to conventional federated learning benchmarks. The empirical evidence illustrates that Fed-RAA considerably diminishes the wall-clock time to reach benchmark accuracies on the MNIST, CIFAR-10, and CIFAR-100 datasets. It notably completes training in 91.06 seconds for MNIST, 89.85 seconds for CIFAR-10, and 101.86 seconds for CIFAR-100, thereby outperforming the closest competitor, FedSync, by over a twofold improvement in speed.

Although Fed-RAA displays slightly reduced accuracy in comparison to top-performing algorithms like RamFed, the marginal accuracy trade-off is counterbalanced by the dramatic reductions in training time, making Fed-RAA an appealing option in time-critical circumstances. This delicate balance of time and accuracy underscores Fed-RAA's practical relevance for rapid model updates and accelerates deployment, a significant contribution to the domain of federated learning that warrants emphasis in our experimental analysis.

It is noteworthy, however, that Fed-RAA's accelerated processing did necessitate a higher number of epochs, with 98399 for MNIST, 98499 for CIFAR-10, and 98599 for CIFAR-100, surpassing the epoch requirements of other baselines. This observation, notwithstanding the reduced time per epoch, implies that Fed-RAA capitalizes on computational efficiency, permitting swift iterative processes even when individual epochs are of minimal computational cost. This attribute makes Fed-RAA particularly valuable in contexts where communication latency poses significant challenges, highlighting the algorithm's capacity to enable rapid convergence and frequent model updates within distributed ML frameworks. Consequently, this dual analysis of time and epoch dynamics bolsters the case for Fed-RAA's adoption in environments where both expedited learning and resource optimization are paramount. 

The experiment details shown in Figure~\ref{fig:exp1_detail} demonstrate the efficacy of the Fed-RAA algorithm in federated learning environments, particularly under asynchronous operations. In the MNIST and CIFAR-10 datasets, Fed-RAA shows significant advantages in balancing accuracy and computational time, outperforming other algorithms such as Fed-Avg, Fed-Sync, and Fed-Prox, which either sacrifice accuracy for lower time costs or vice versa. For the more complex CIFAR-100 dataset, Fed-RAA remarkably excels by achieving higher accuracies more rapidly compared to its counterparts, which face sharp increases in computational time with minimal accuracy gains. This consistent performance underscores the benefits of asynchronous learning and training submodels, suggesting that Fed-RAA's approach to handling partial model training can substantially enhance the efficiency and effectiveness of federated learning systems.
\begin{table}[ht!]
\caption{Comparison of accuracy and time cost across different baselines}
\label{comparison_across_datasets}
\centering
\vspace{\baselineskip}
\begin{tabular}{lcccccc}
\toprule
Baseline & \multicolumn{2}{c}{MNIST} & \multicolumn{2}{c}{CIFAR-10} & \multicolumn{2}{c}{CIFAR-100} \\
\cmidrule(r){2-3} \cmidrule(r){4-5} \cmidrule(r){6-7}
 & Accuracy (\%) & Time (s) & Accuracy (\%) & Time (s) & Accuracy (\%) & Time (s) \\
\midrule
FedAvg   & 96.35 & 1219.05 & 52.88 & 1172.49 & 25.22 & 1306.53 \\
FedSync  & 96.50 &  727.38  & 53.01 & 759.33  & 25.41 & 3369.75  \\
FedProx  & 96.40 & 2599.20 & 52.89 &  2611.32 & 25.87 & 2732.94 \\
FedPrun  & 94.13 & 2330.46 & 53.60 & 2926.44 & 24.97 & 9475.65 \\
SplitFL  & 95.30 & 2520.45 & 51.30 & 2309.64 & 23.68 &  6544.77 \\
Ram-Fed   & 96.55 &  2426.88 & 54.22 & 2349.18 & 27.04 & 2754.33\\
Fed-RAA   & 93.86 & 273.18  & 47.38 & 269.54 & 20.55 & 305.58  \\
\bottomrule
\end{tabular}
\end{table}

\subsection{Detailed Results of Comparison for \(beta\) and \(M\)}
\begin{figure}[ht!]
\centering
% Row 1: MNIST
\begin{subfigure}[b]{0.22\textwidth}
    \includegraphics[width=\textwidth]{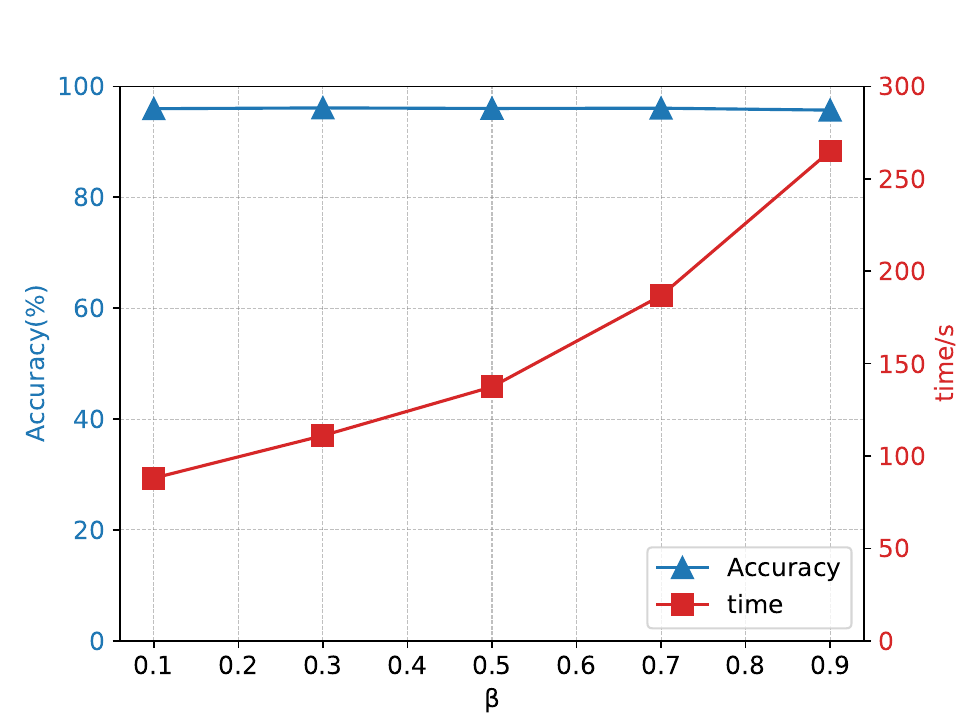}
    \caption{\(M = 2\)}
\end{subfigure}
\hfill
\begin{subfigure}[b]{0.22\textwidth}
    \includegraphics[width=\textwidth]{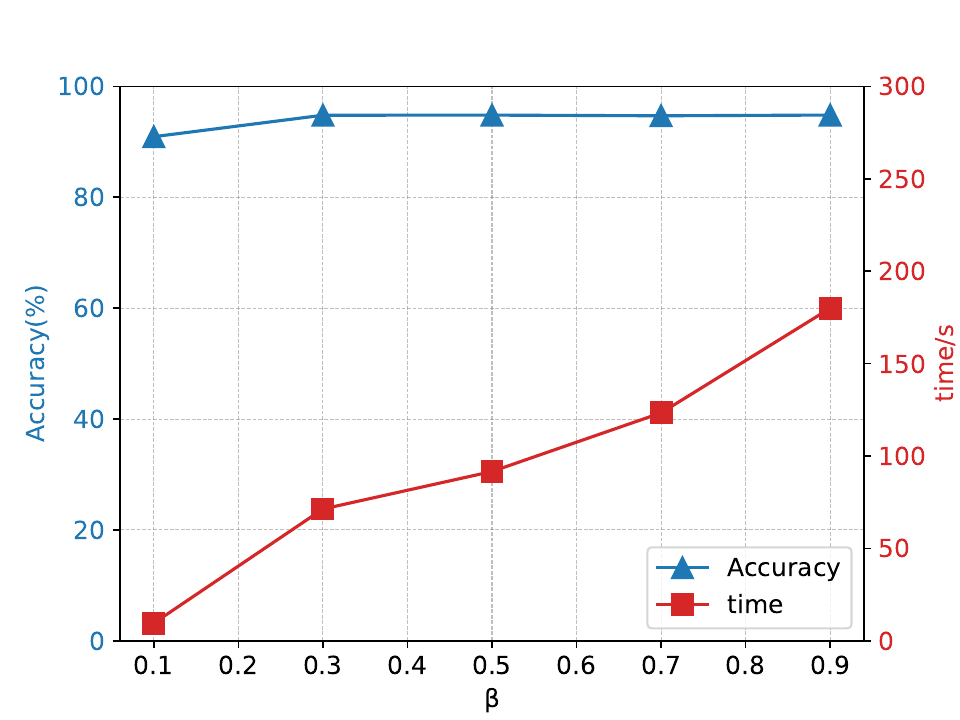}
    \caption{\(M = 3\)}
\end{subfigure}
\hfill
\begin{subfigure}[b]{0.22\textwidth}
    \includegraphics[width=\textwidth]{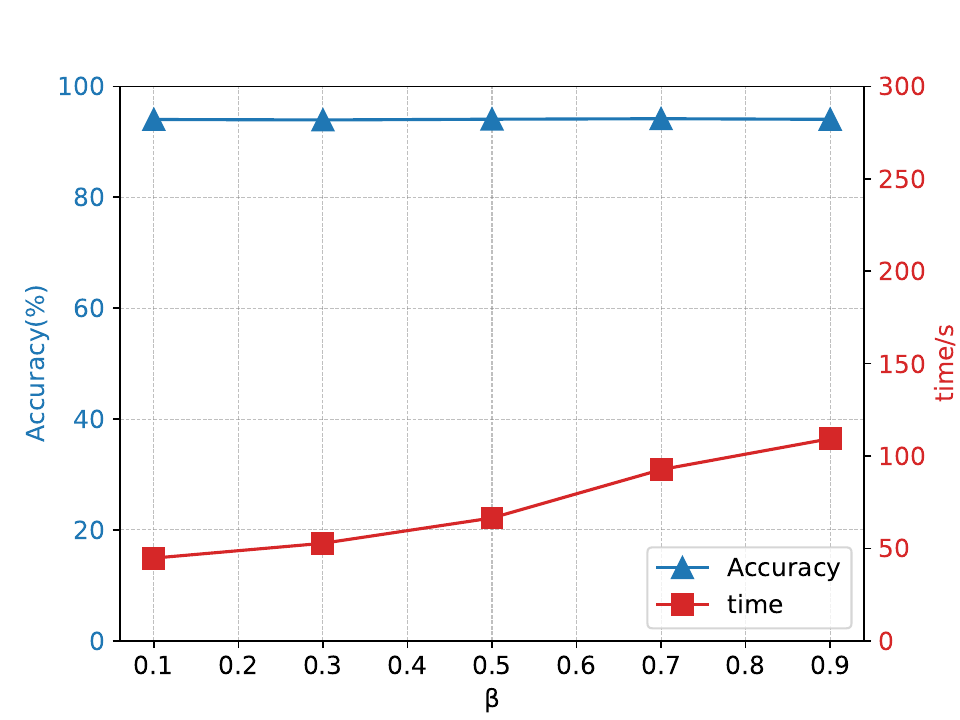}
    \caption{\(M = 4\)}
\end{subfigure}
\hfill
\begin{subfigure}[b]{0.22\textwidth}
    \includegraphics[width=\textwidth]{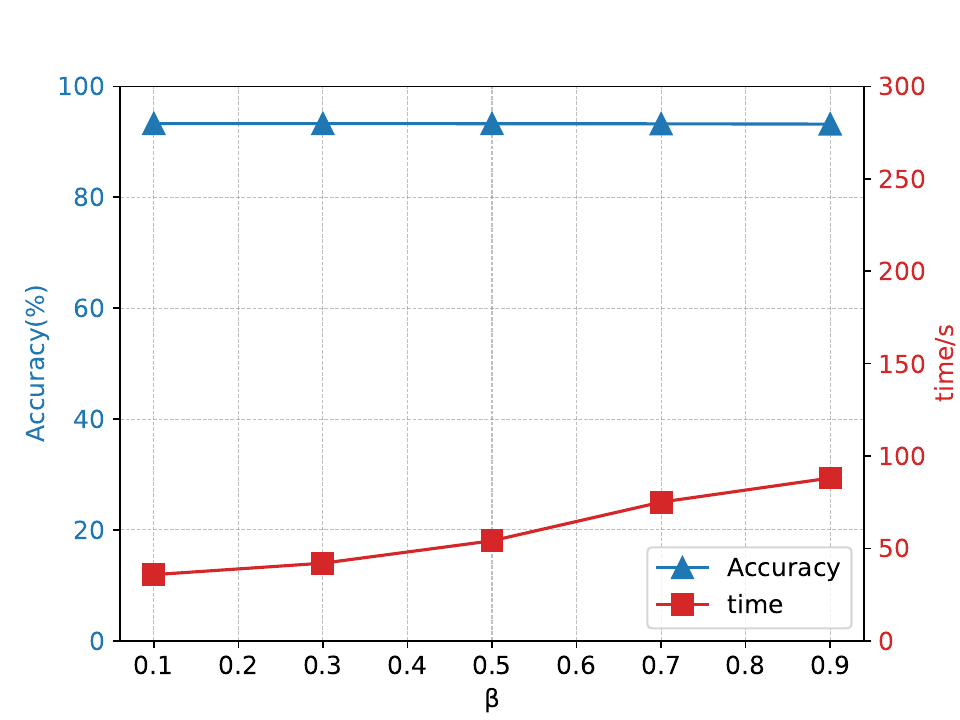}
    \caption{\(M = 5\)}
\end{subfigure}
\caption{Results of the MNIST dataset.}

\vspace{1em} % Adds some space between the rows

% Row 2: CIFAR-10
\begin{subfigure}[b]{0.22\textwidth}
    \includegraphics[width=\textwidth]{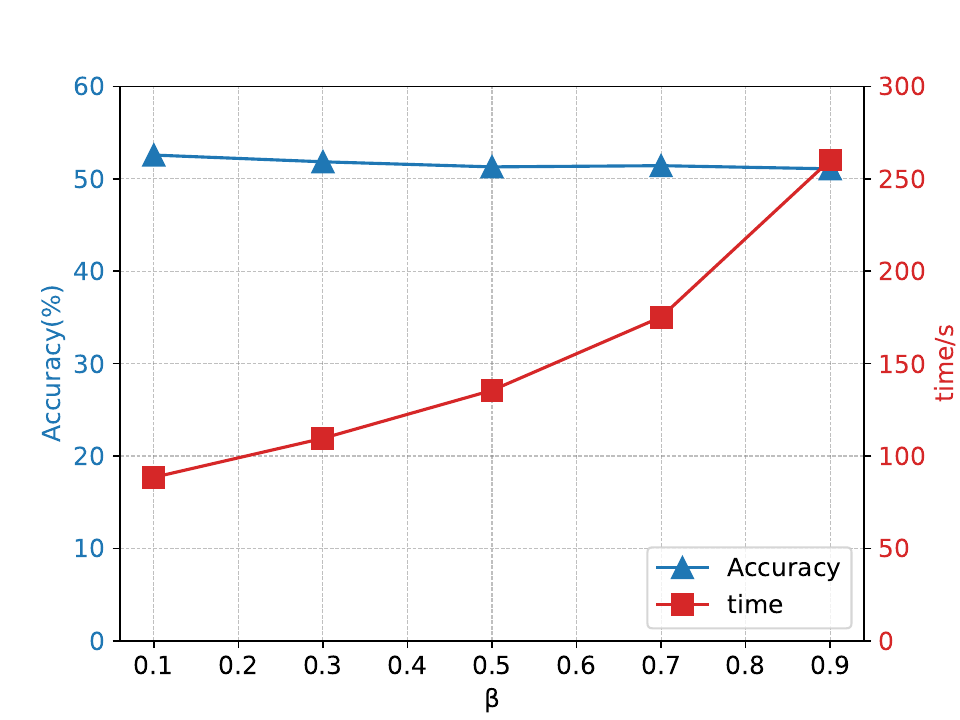}
    \caption{\(M = 2\)}
\end{subfigure}
\hfill
\begin{subfigure}[b]{0.22\textwidth}
    \includegraphics[width=\textwidth]{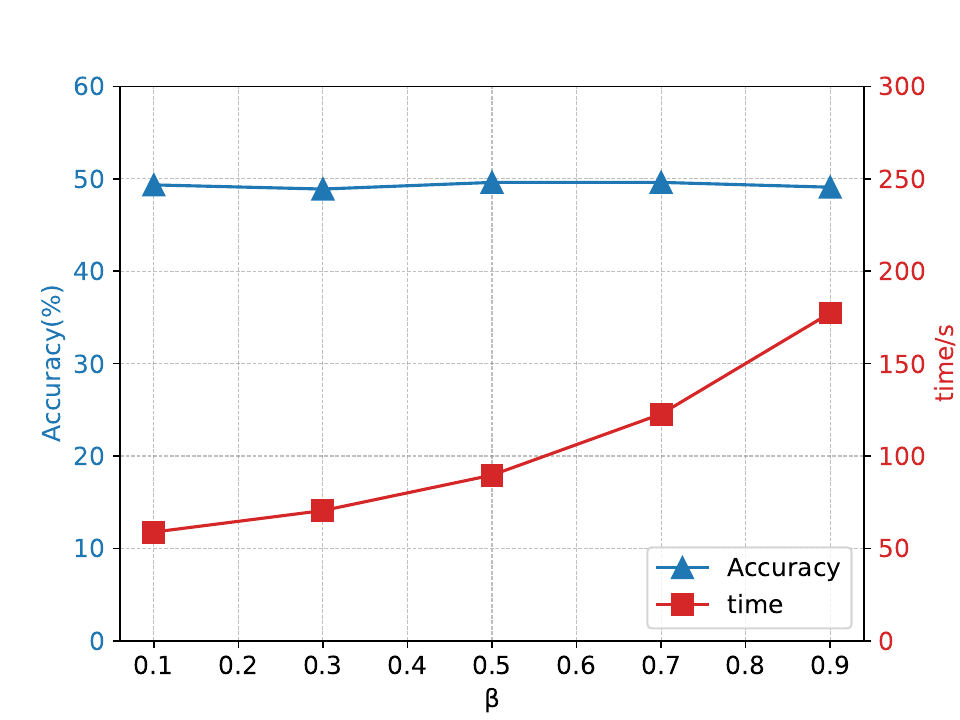}
    \caption{\(M = 3\)}
\end{subfigure}
\hfill
\begin{subfigure}[b]{0.22\textwidth}
    \includegraphics[width=\textwidth]{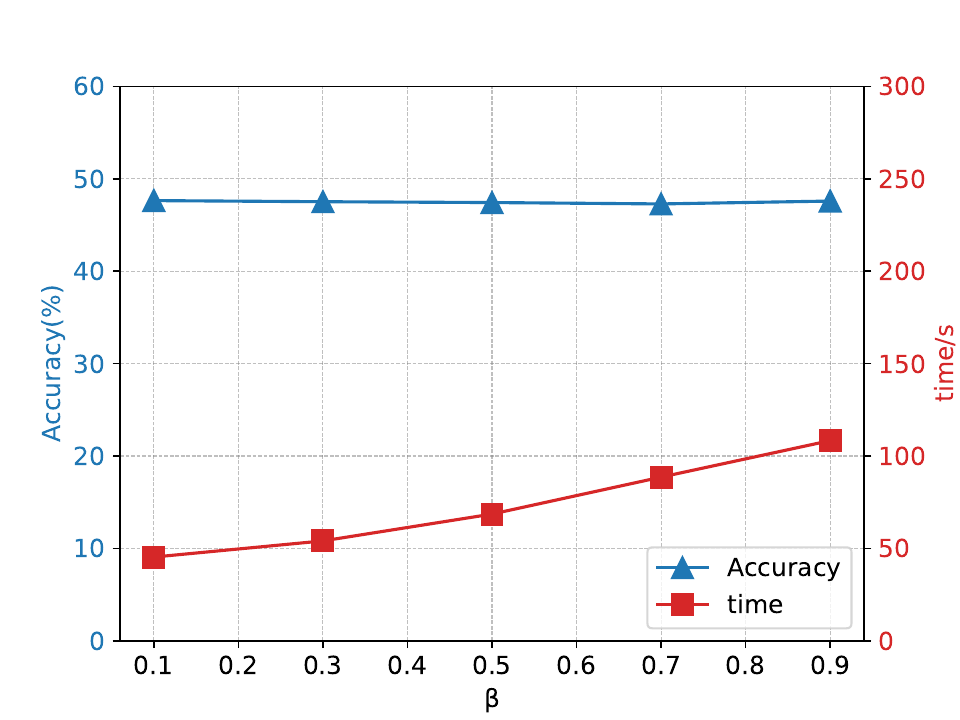}
    \caption{\(M = 4\)}
\end{subfigure}
\hfill
\begin{subfigure}[b]{0.22\textwidth}
    \includegraphics[width=\textwidth]{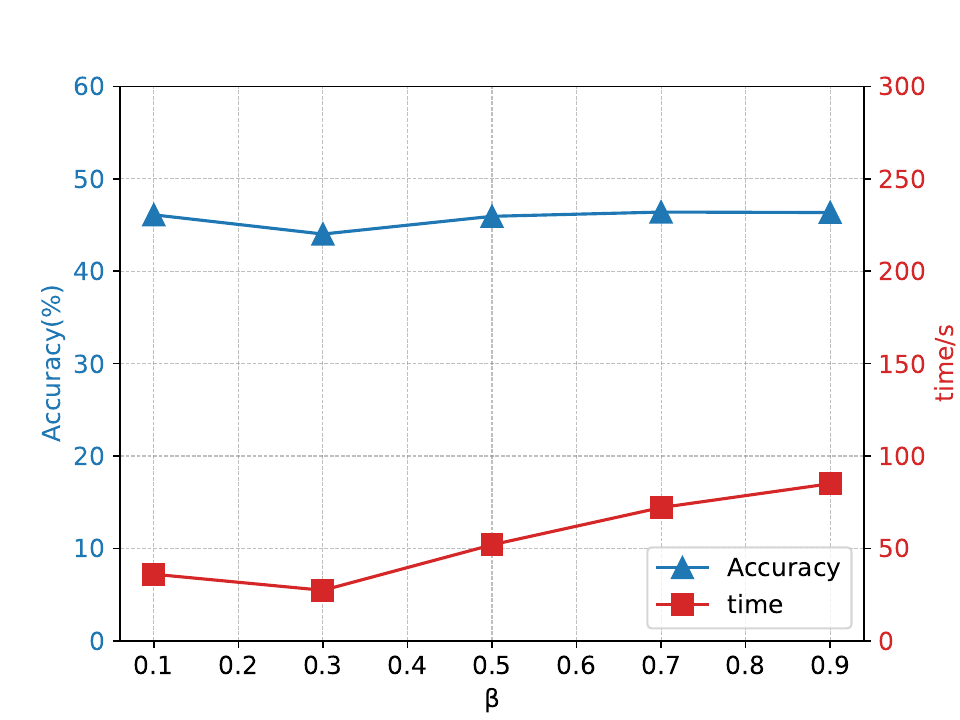}
    \caption{\(M = 5\)}
\end{subfigure}
\caption{Results of the CIFAR-10 dataset.}

\vspace{1em} % Adds some space between the rows

% Row 3: CIFAR-100
\begin{subfigure}[b]{0.22\textwidth}
    \includegraphics[width=\textwidth]{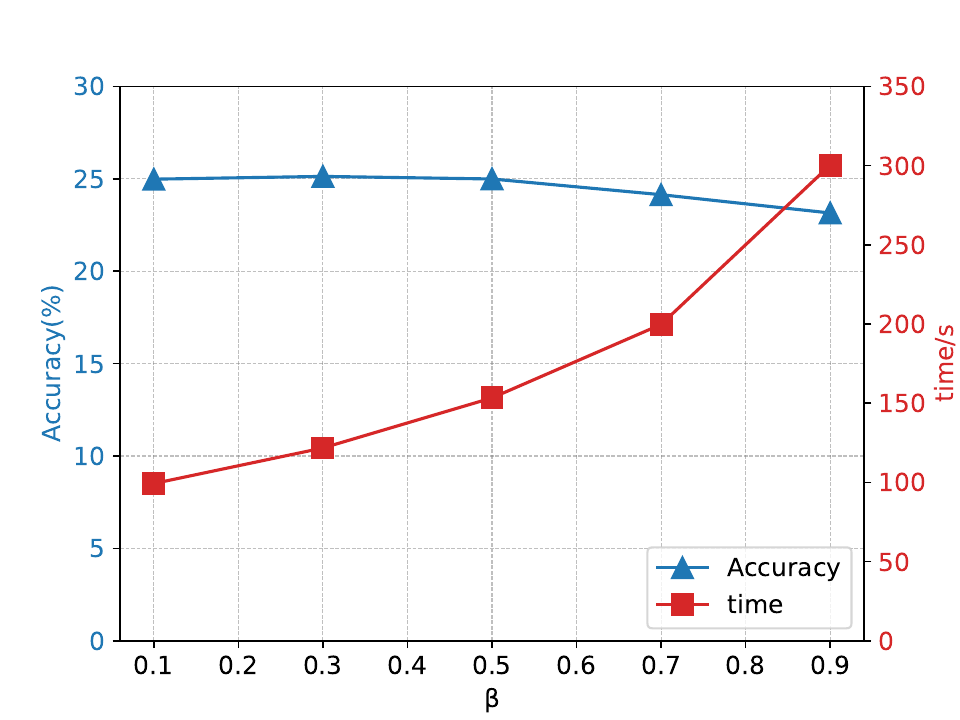}
    \caption{\(M = 2\)}
\end{subfigure}
\hfill
\begin{subfigure}[b]{0.22\textwidth}
    \includegraphics[width=\textwidth]{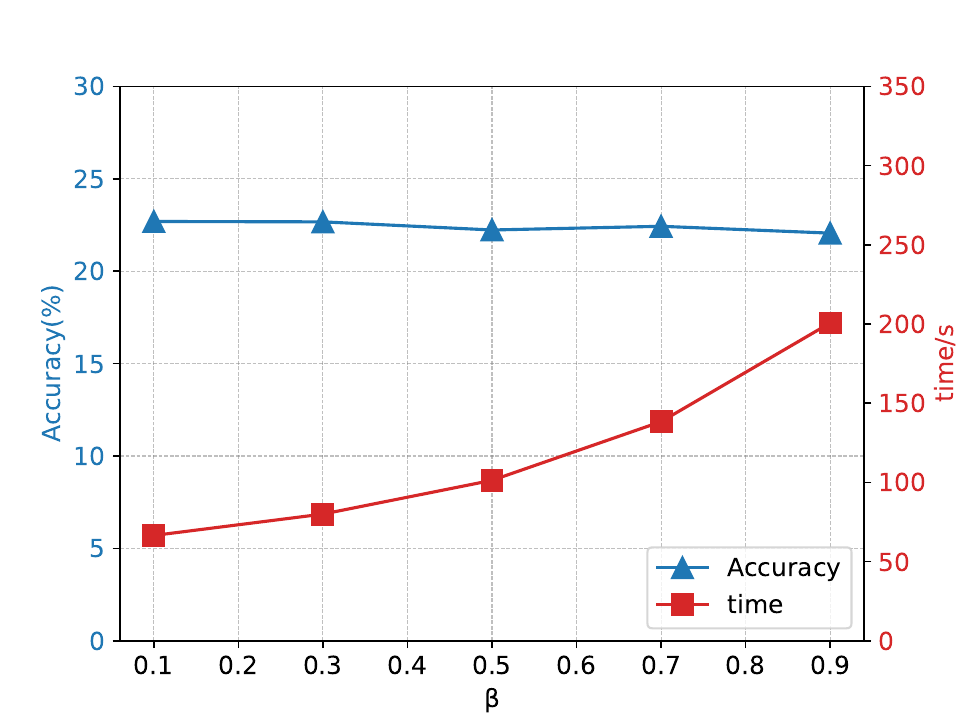}
    \caption{\(M = 3\)}
\end{subfigure}
\hfill
\begin{subfigure}[b]{0.22\textwidth}
    \includegraphics[width=\textwidth]{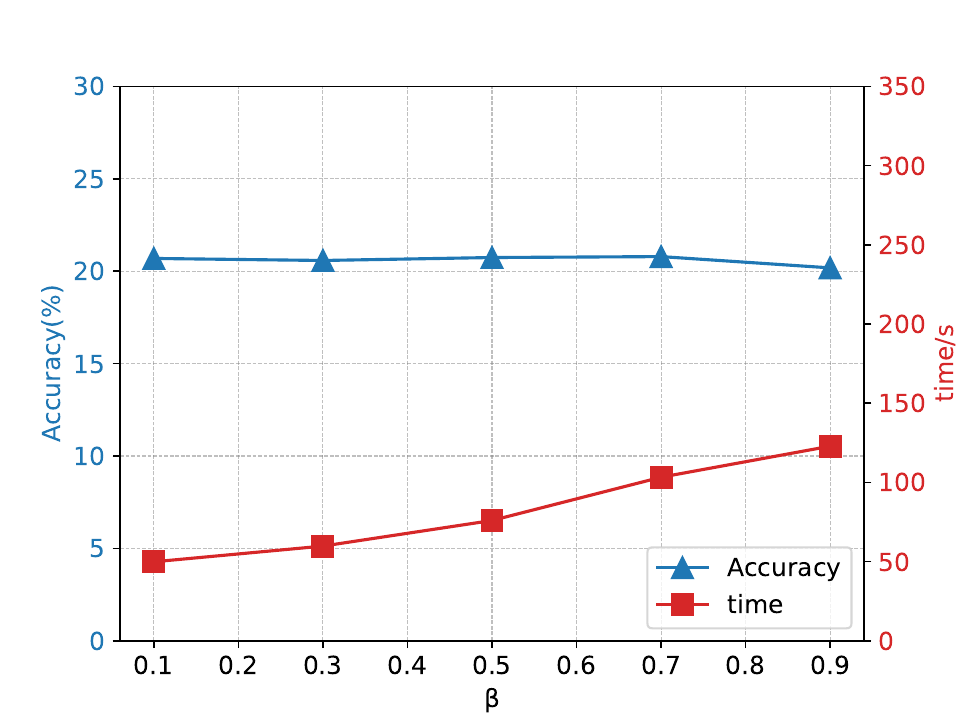}
    \caption{\(M = 4\)}
\end{subfigure}
\hfill
\begin{subfigure}[b]{0.22\textwidth}
    \includegraphics[width=\textwidth]{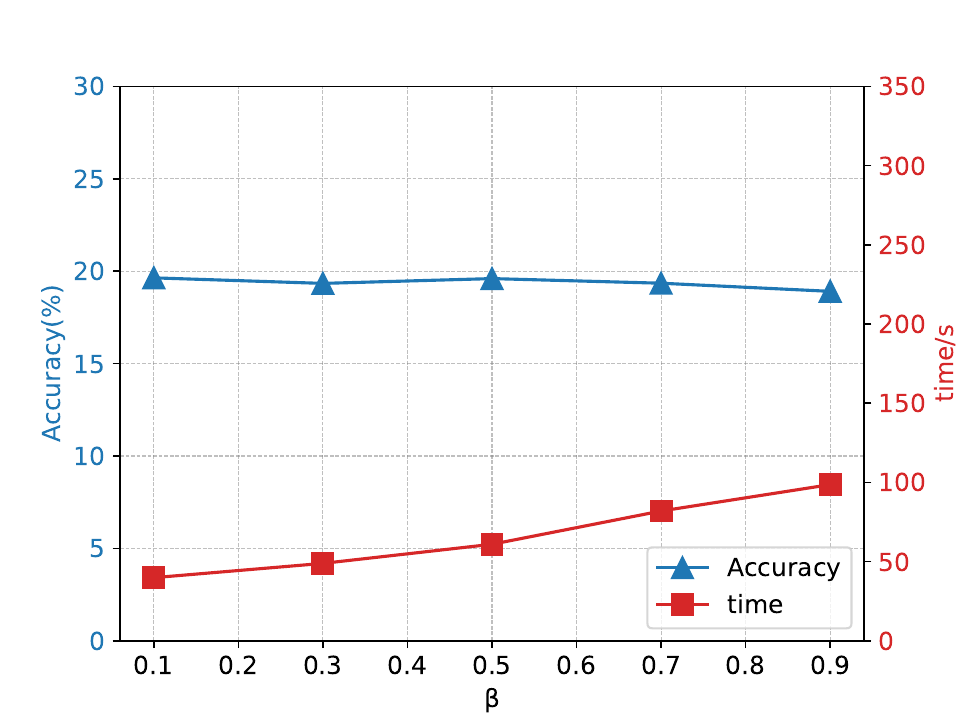}
    \caption{\(M = 5\)}
\end{subfigure}
\caption{Results of the CIFAR-100 dataset.}

\caption{Detailed results of comparison for \(\beta\) of different submodel numbers \(M\) on dataset MNIST, CIFAR-10, and CIFAR-100.}
\label{detailed_beta_M}
\end{figure}
In the experimental analyses depicted in Figure~\ref{fig:beta}, we observed that increasing the proportion of weak-capability clients (\(\beta\)) significantly impacts both the training time and accuracy of federated learning models across various datasets. 
Specifically, as \(\beta\) was raised, particularly to 0.9, a clear increase in training time was noted across all datasets, with the most pronounced effects in the CIFAR-100 dataset due to its higher complexity. 
This indicates that a higher number of less capable clients markedly slows down the learning process. 
Accuracy trends also varied with dataset complexity; 
the MNIST dataset maintained relative stability in accuracy until a sharp decline at \(\beta = 0.9\), whereas both CIFAR datasets exhibited a more consistent and sensitive decline in accuracy as \(\beta\) increased. 
These observations underscore the significant influence of client capability distribution on the performance of federated learning systems, particularly in complex tasks. 
The findings highlight the crucial need for advanced mechanisms to manage client heterogeneity, suggesting that adaptive strategies and careful management of client capabilities are essential for enhancing both the efficiency and effectiveness of federated learning outcomes in diverse and challenging data environments.

Results shown in Figure~\ref{fig:M} reveal significant insights into the impact of model fragmentation on training dynamics. 
Notably, increasing the number of submodels from 2 to 5 consistently led to prolonged training times, with the effect being most pronounced in the CIFAR-100 dataset. 
This dataset, characterized by its complexity, showed steep increases in training duration without corresponding improvements in accuracy, suggesting a scalability challenge with Fed-RAA when managing numerous small submodels in complex environments. 
Moreover, while the MNIST and CIFAR-10 datasets exhibited more moderate increases in training time, the benefit in accuracy did not substantially improve with more submodels, particularly in CIFAR-10 where training times converged across submodel configurations as accuracy increased. These observations indicate that there exists a critical trade-off in the asynchronous federated learning context of Fed-RAA between the number of submodels and the efficiency of training. Smaller submodels, although potentially beneficial for individual client training cycles, may introduce significant overhead in managing multiple updates, especially in datasets with higher complexity. This underscores the necessity for optimizing the balance between model fragmentation and training efficiency to enhance the effectiveness of federated learning systems in diverse data conditions.

\section{Discussion}
\textbf{Limitations.}
Although our proposed algorithm, Fed-RAA, serves as a foundational approach that is adaptable to any model partitioning strategy, it is important to recognize that its effectiveness could be significantly enhanced by tailoring the partitioning strategy to account for the heterogeneous and constrained resources at the client level. This consideration emerges as a limitation in the current design of Fed-RAA, where the generic approach to model fragmentation does not explicitly address variations in client capabilities. By integrating a partitioning mechanism that dynamically adjusts to the resource constraints and computational capabilities of individual clients, the performance of Fed-RAA could be optimized, leading to improved efficiency and effectiveness in practical federated learning scenarios. This potential refinement highlights a valuable direction for future enhancements and underscores the importance of context-aware strategies in the development of federated learning algorithms.
In addition, we refer to the following articles~\cite{zhang2024collaborative,zou2024distributed,zhang2024modeling,gao2024cooperative}

\textbf{Impacts. }
The algorithm introduced in this paper, Fed-RAA, has meaningful positive impacts on the field of federated learning. It empowers resource-constrained edge network devices to collaboratively train and deploy large models, thereby democratizing access to sophisticated machine learning capabilities which would be impractical with traditional centralized training methods. Moreover, as a foundational and neutral algorithm, Fed-RAA avoids inherent biases and specific adverse societal impacts. This characteristic makes it an adaptable framework that can be tailored by researchers and practitioners to align with ethical standards and societal needs, ensuring a conscientious deployment of AI technologies that mitigates risks associated with more specialized algorithms potentially amplifying biases or fostering inequality.

\end{document}